\documentclass{article}

\usepackage{times}
\usepackage{natbib}
\usepackage{marvosym}
\setcitestyle{authoryear,round,citesep={;},aysep={,},yysep={;}}

\makeatletter
\renewcommand{\paragraph}{\@startsection{paragraph}{4}{\z@}{1.0ex plus 0.3ex minus 0.1ex}{-1em}{\normalfont\normalsize\bfseries}}
\makeatother

\usepackage{amsmath,amsfonts,bm}

\def\eqref#1{equation~\ref{#1}}

\def\1{\bm{1}}

\def\va{{\bm{a}}}

\def\vo{{\bm{o}}}

\def\vr{{\bm{r}}}

\def\vu{{\bm{u}}}

\def\vx{{\bm{x}}}

\DeclareMathAlphabet{\mathsfit}{\encodingdefault}{\sfdefault}{m}{sl}
\SetMathAlphabet{\mathsfit}{bold}{\encodingdefault}{\sfdefault}{bx}{n}

\usepackage{amsmath}
\usepackage{amssymb}
\usepackage{mathtools}
\usepackage{graphicx}
\usepackage{booktabs}
\usepackage{multirow}
\usepackage{tabularx}
\usepackage{array}
\usepackage{xcolor}
\usepackage{xspace}
\usepackage{enumitem}
\usepackage{subcaption}
\DeclareCaptionFont{calibrationsmall}{\fontsize{5}{6}\selectfont}
\usepackage{wrapfig}
\usepackage{microtype}
\usepackage{url}

\usepackage{hyperref}
\usepackage[nameinlink,capitalise,noabbrev]{cleveref}
\crefname{section}{Section}{Sections}
\Crefname{section}{Section}{Sections}
\crefname{figure}{Figure}{Figures}
\Crefname{figure}{Figure}{Figures}
\crefname{table}{Table}{Tables}
\Crefname{table}{Table}{Tables}
\crefname{equation}{Equation}{Equations}
\Crefname{equation}{Equation}{Equations}
\hypersetup{
  colorlinks=true,
  linkcolor=blue!55!black,
  citecolor=green!40!black,
  urlcolor=blue!60!black
}

\newcommand{\method}{\textsc{DreamTrue}\xspace}
\newcommand{\plucker}{Pl\"ucker\xspace}

\newcommand{\ndtw}{nDTW\xspace}

\definecolor{rankblue}{HTML}{4088D6}
\newcommand{\rankbest}[1]{\begingroup\setlength{\fboxsep}{1.2pt}\colorbox{rankblue!36}{\strut #1}\endgroup}
\newcommand{\ranksecond}[1]{\begingroup\setlength{\fboxsep}{1.2pt}\colorbox{rankblue!20}{\strut #1}\endgroup}
\newcommand{\rankthird}[1]{\begingroup\setlength{\fboxsep}{1.2pt}\colorbox{rankblue!10}{\strut #1}\endgroup}

\newif\ifdraftnotes
\draftnotestrue

\newcolumntype{Y}{>{\centering\arraybackslash}X}
\newcolumntype{L}[1]{>{\raggedright\arraybackslash}p{#1}}
\newcolumntype{C}[1]{>{\centering\arraybackslash}p{#1}}

\setlist[itemize]{leftmargin=*,topsep=2pt,itemsep=1pt}
\setlist[enumerate]{leftmargin=*,topsep=2pt,itemsep=1pt}

\title{DreamTrue: Action-Faithful Robot World Model with Counterfactual Post-Training}
\providecommand{\authorskipshort}{\hspace{0.5em}}
\providecommand{\Letter}{\ensuremath{\dagger}}
\author{
  \begin{tabular}{c}
  \textbf{Junyan~Li}\textsuperscript{1,2*} \authorskipshort
  \textbf{Ruizhi~Li}\textsuperscript{1,2*} \authorskipshort
  \textbf{Yu~Liu}\textsuperscript{2\dag} \authorskipshort
  \textbf{Xiangshuo~Liu}\textsuperscript{1\S} \authorskipshort
  \\
  \textbf{Mingchao~Sun}$^{2}$ \authorskipshort
  \textbf{Hongyu~Pan}$^{2}$ \authorskipshort
  \textbf{Mu~Xu}$^{2}$ \authorskipshort
  \textbf{Lue~Fan}\textsuperscript{1\dag\Letter} \authorskipshort
  \textbf{Zhaoxiang~Zhang}\textsuperscript{1\Letter}
  \end{tabular}
  \\
  $^1$NLPR, Institute of Automation, Chinese Academy of Sciences (CASIA) \\
  $^2$Amap, Alibaba Group\\[1pt]
  {\small $^*$ Equal contribution. \quad
  \textsuperscript{\dag} Co-Project Leads.}\\[1pt]
  {\small\texttt{
  \{lue.fan, zhaoxiang.zhang\}@ia.ac.cn
  }}
}
\date{}

\begin{document}

\maketitle
\begingroup
\renewcommand{\thefootnote}{\S}
\footnotetext{Xiangshuo Liu was an intern at CASIA during this work.}
\endgroup
\begin{abstract}
We present \method, a multi-view, cross-embodiment
robot world model for action-faithful and physically
plausible video prediction.
Training such a model on existing robot datasets faces
two obstacles: imprecise calibration can impair action
following, while limited coverage of unsuccessful
interactions can bias predictions toward successful
outcomes.
To improve action following across embodiments, we render
action trajectories into image-space conditions and
introduce offline geometric calibration to align
these conditions with the target videos.
To broaden interaction coverage, we introduce
counterfactual post-training, modifying recorded
action trajectories and generating future videos
under a wider range of actions and contact configurations.
To provide feedback on these predictions without paired
ground-truth futures, we construct a human-annotated
video dataset covering robot, object, and interaction
defects and use it to train an embodied video reward model.
Its scores guide reinforcement-learning post-training
toward more physically plausible interaction outcomes.
On AgiBot, \method attains state-of-the-art action following,
while reducing the human-assessed interaction defect rate from
from 48.12\% to 6.25\%.
Notably, our model ranks first in the world model track
of the AgiBot World Challenge 2026.
The project page can be found at
\url{https://brave-eai.github.io/DreamTrue}.

\end{abstract}

\section{Introduction}
\label{sec:introduction}

\begin{figure}[htbp]
    \centering
    \includegraphics[width=0.95\linewidth]{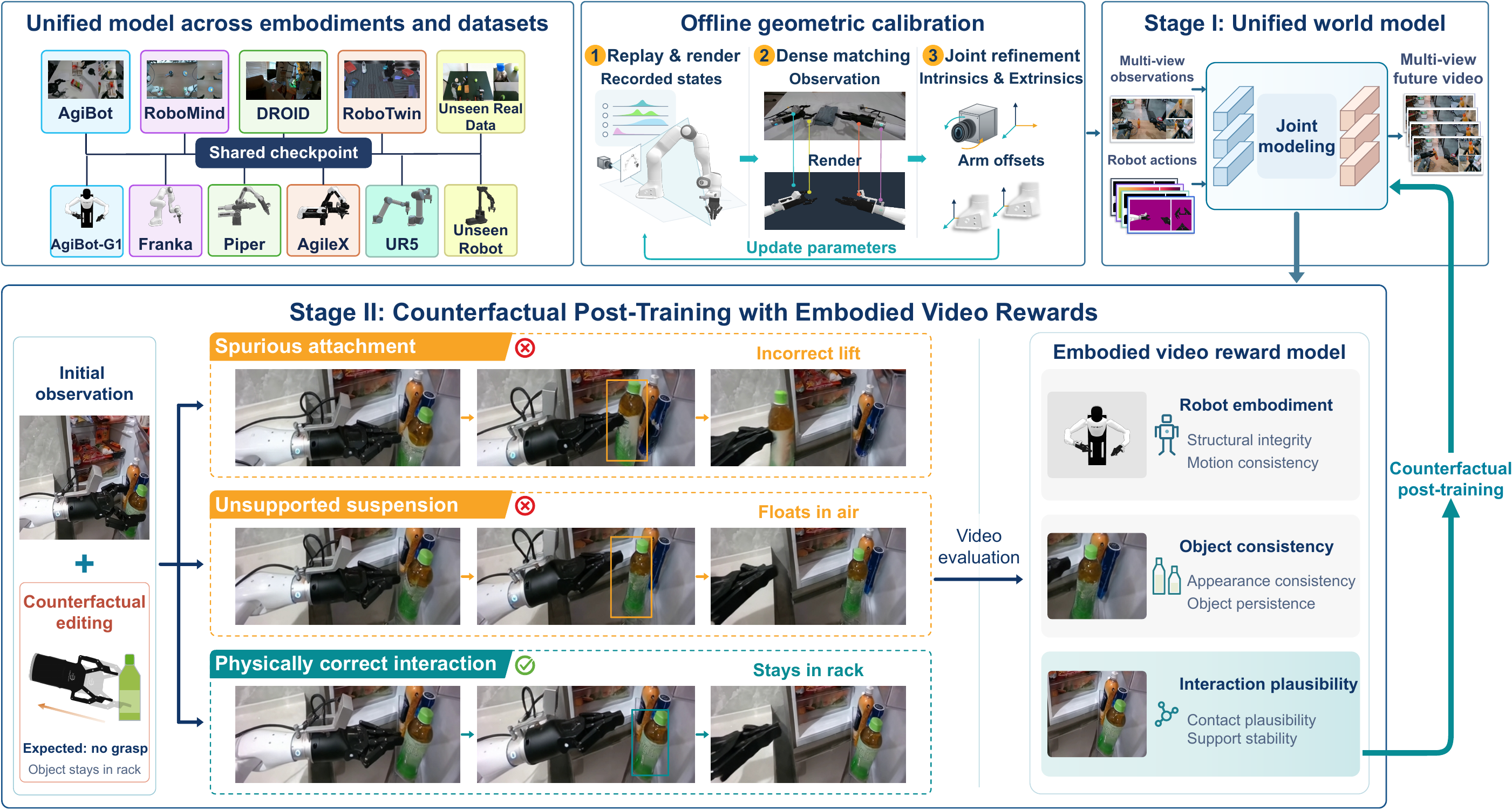}
    \caption{Overview of \method. Camera calibration aligns robot renderings with recorded
videos. Stage I learns action-conditioned prediction
across embodiments, while Stage II uses video rewards
for counterfactual post-training to improve interaction
plausibility.} \label{fig:method-overview}
    \vspace{-20pt}
\end{figure}

Robot world models predict how a scene evolves under a given action sequence, providing virtual environments for robot learning and evaluation~\citep{ha_recurrent_2018, hafner_dreamer_2020, zhou_robodreamer_2024, li_comprehensive_2025}. To be reliable as simulators, they must accurately follow the specified actions and produce physically plausible interactions with surrounding objects~\citep{yue_ewmbench_2025, shang_worldarena_2026}.

Public robot datasets provide extensive action--video
recordings for learning these capabilities~\citep{openx_embodiment_2024,khazatsky_droid_2024,wu2025robomind,bu2025agibot_iros}, but training
from these data faces two challenges.
First, action conditions must accurately represent
the robot's intended trajectory in the target video.
Rendering actions through each robot's model provides
direct visual guidance for action following by expressing
different embodiments' actions in the image space of
the target video~\citep{chen_bridgev2w_2026,
alzayer_masked_2026}.
This approach, however, depends on accurate camera
calibration, which is not always available in existing
recordings.
Calibration errors can spatially misalign the rendered
robot with its appearance in the recorded video,
making the action conditions inconsistent with
the training targets.
Second, training data dominated by successful
demonstrations provide limited coverage of missed
grasps, slipping objects, and unsuccessful contact~\citep{peng_fact_2026}.
Recent work identifies a tendency to predict successful
outcomes even under actions that lead to
failure.
For example, a model may predict that an object
rises with the gripper despite a missed grasp.
Such hallucinations can overestimate
action success, misleading action selection and
producing overly optimistic policy evaluations.

Collecting new real-robot trajectories with accurate
calibration and broad coverage of failed interactions
would help address these challenges, but would require
substantial time and resources~\citep{khazatsky_droid_2024}.
We therefore seek to improve the quality of existing
data by correcting calibration errors and broaden
model training by constructing counterfactual action
trajectories, without additional real-robot data
collection.

To this end, we introduce \method, a multi-view,
cross-embodiment robot world model
for action-faithful
and physically plausible video prediction.
To improve the accuracy of action--video supervision,
we propose \textbf{offline geometric calibration}
to align robot renderings with recorded video frames.
Known robot geometry and recorded robot states allow
us to render the robot and establish image
correspondences with the observations.
These correspondences guide camera parameter refinement
without dedicated calibration sequences.
The optimized parameters are then used to render action
trajectories, providing spatially aligned visual
conditions for supervised training across robot datasets.
To broaden the actions and interactions encountered
during training, we propose \textbf{counterfactual
post-training with embodied video rewards}.
For a given scene, we modify recorded 
trajectories to construct counterfactual actions that
may lead to unsuccessful interactions, and condition
the world model on these actions to generate future
videos.
Although paired ground-truth futures are unavailable,
some prediction errors remain directly observable:
an object rising after a missed grasp or remaining
suspended after losing support reveals an implausible
interaction (\Cref{fig:method-overview}).
We therefore construct a human-annotated video defect
dataset covering videos generated by multiple world
models.
Annotations span robot embodiment, object consistency,
and interaction plausibility, and are used to train
an embodied video reward model.
Its scores guide reinforcement-learning post-training
to reduce observable defects and encourage physically
plausible interaction outcomes.
This provides feedback under actions beyond the
recorded trajectories without requiring additional
real-world collection of failed executions.

We train \method across multiple public robot datasets and evaluate action following, visual quality, and interaction plausibility. On the AgiBot benchmark, \method achieves the highest nDTW score
of 0.8772 among the evaluated methods for action
following and reduces the human-assessed interaction
defect rate from 48.12\% to 6.25\% through
reward-guided post-training.
Our contributions are threefold:

\begin{itemize}
    \item We introduce \method, \textbf{a multi-view, cross-embodiment robot world model} that improves action following and interaction plausibility without additional real-robot training data collection.
    \item We propose \textbf{an offline geometric calibration method} to improve alignment between action conditions and recorded videos without dedicated calibration sequences. We release the resulting calibration for three datasets, covering 153,666 retained episodes and over 1,660 hours.
    \item We develop \textbf{a reward-guided post-training approach} that learns from human annotations of robot, object, and interaction defects, enabling feedback under counterfactual actions without paired future videos. We release annotations for 44.9K videos, including 30.4K defect annotations.
\end{itemize}

\section{Related Work}
\label{sec:related-work}

\paragraph{Action-conditioned robot world models.}
Action-conditioned world models leverage video-generation priors to predict robot motion and environmental responses~\citep{li_comprehensive_2025,yang_neoverse_2026}. Ctrl-World enables multi-view, long-horizon prediction, while Genie Envisioner and GE-Sim 2.0 use action-conditioned simulation for policy learning and closed-loop evaluation~\citep{guo_ctrl-world_2025,liao_genie_envisioner_2025,qiu_gesim2_2026}. To expand interaction coverage, DreamDojo transfers knowledge from large-scale human videos, PlayWorld uses autonomous robot interaction data, and FACT predicts future videos and task progress from failure trajectories~\citep{gao_dreamdojo_2026,yin_playworld_2026,peng_fact_2026}. Learned world-model rollouts support policy evaluation, data generation, and policy optimization~\citep{team_evaluating_2025,nvidia_world_2025,zhu_wmpo_2025,guo_vlaw_2026,jiang_wovr_2026,sun_atomvla_2026}. Our model follows this line while emphasizing action-faithful and physically coherent predictions under counterfactual actions.

\paragraph{Geometric action representations and robot--camera calibration.}
BridgeV2W renders robot states into pixel-aligned embodiment masks, while Masked Visual Actions constructs visual action conditions through segmentation or robot rendering~\citep{chen_bridgev2w_2026,alzayer_masked_2026}; EnerVerse-AC further adds camera-ray encoding for multi-view generation~\citep{jiang_enerverse_ac_2025}. Their fidelity depends on accurate robot--camera geometry. Classical hand--eye calibration solves the $AX=XB$ relation, while markerless methods use differentiable rendering; both require dedicated calibration sequences~\citep{tsai_lenz_1989,chen_easyhec_2023,hong_easyhecpp_2024}. PointWorld instead refines camera poses from existing recordings but relies on accurate depth from sensors or FoundationStereo~\citep{huang_pointworld_2026,wen_foundationstereo_2025}. We recover camera parameters and robot mounting offsets from RGB demonstrations alone, without dedicated sequences or depth, making the approach applicable to most open manipulation datasets.

\paragraph{Embodied video evaluation and reward-based post-training.}
EWMBench evaluates embodied videos by scene, motion, and semantics, while WorldArena combines perceptual quality with downstream functional evaluation~\citep{yue_ewmbench_2025,shang_worldarena_2026}. WorldCompass post-trains interactive world models without paired futures, using rewards for camera-trajectory following and frame-level visual quality~\citep{wang_worldcompass_2026,li_pisa_2025,yuan_inference-time_2026,ping_flow-factory_2026}. These rewards primarily capture agent motion and generic visual quality rather than physical object interaction. We instead learn an embodied video reward from human annotations of robot, object, and interaction defects and use it to post-train physically coherent object interaction under counterfactual robot actions.

\section{Method}
\label{sec:method}

\subsection{Problem Formulation and Overview}
\label{sec:method-overview}

Given initial multi-view observations $\vx_0^{1:V}$,
a task instruction $\ell$, and an action sequence
$\va_{1:T}$, we model the distribution of future videos as
\begin{equation}
\hat{\vx}_{1:T}^{1:V}
\sim
p_\theta\!\left(
\vx_{1:T}^{1:V}
\mid
\vx_0^{1:V},\ell,\va_{1:T}
\right),
\label{eq:world-model}
\end{equation}
where $V$ is the number of camera views, $T$ is the prediction horizon, and $\theta$ denotes model parameters. Our framework consists of offline geometric calibration and two training stages. We first refine camera parameters and optional robot mounting offsets to improve spatial alignment between robot renderings and recorded video frames. Stage I represents action sequences as robot renderings and trains a multi-view generator across embodiments for accurate action following. Stage II uses a learned video reward model to guide post-training toward more plausible robot–object interactions.

\subsection{Offline Geometric Calibration}
\label{sec:camera-refinement}

We refine camera parameters using image correspondences from existing recordings to improve alignment between robot renderings and video frames, without dedicated calibration sequences.

\paragraph{Image correspondence construction.}

\begin{wrapfigure}{r}{0.40\textwidth}
    \vspace{-10pt}
    \centering
    \captionsetup{font=footnotesize}
    \begin{subfigure}[t]{0.32\linewidth}
        \centering
        \includegraphics[width=\linewidth]{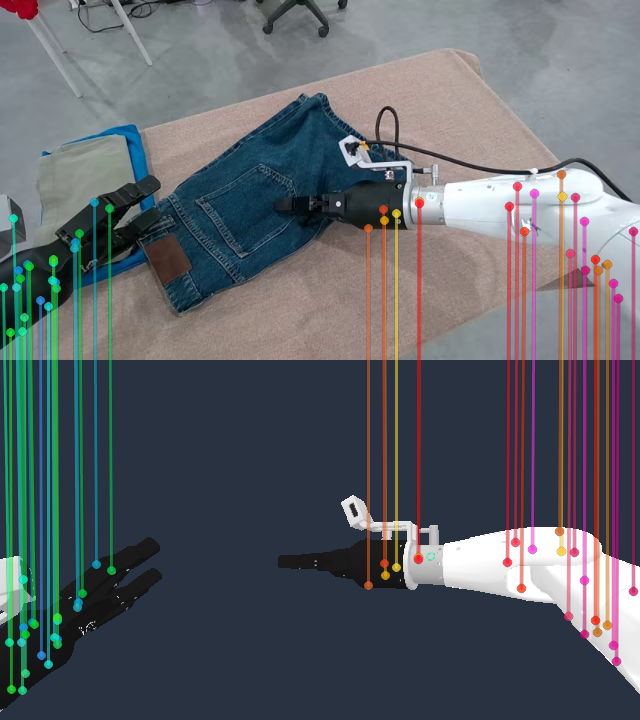}
        \caption{$\mathcal{P}_{\mathrm{align}}$}
        \label{fig:calibration-align}
    \end{subfigure}
    \begin{subfigure}[t]{0.32\linewidth}
        \centering
        \includegraphics[width=\linewidth]{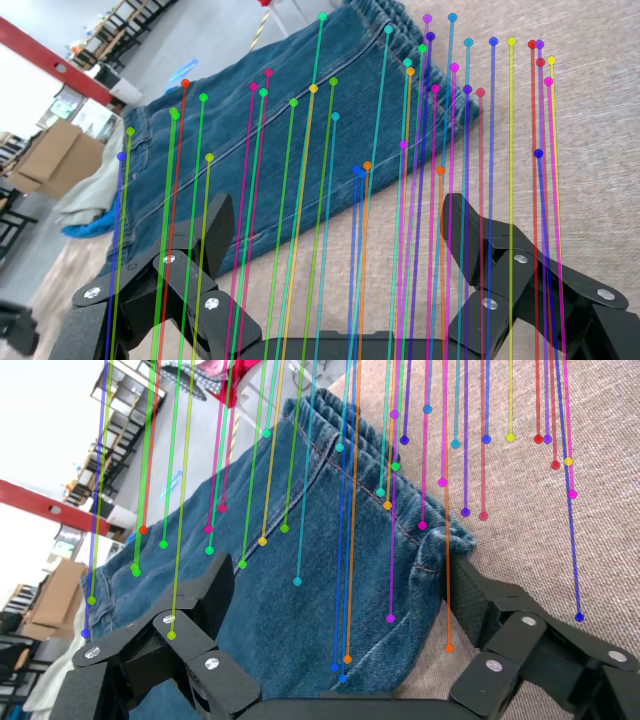}
        \caption{$\mathcal{P}_{\mathrm{time}}$}
        \label{fig:calibration-time}
    \end{subfigure}
    \begin{subfigure}[t]{0.32\linewidth}
        \centering
        \includegraphics[width=\linewidth]{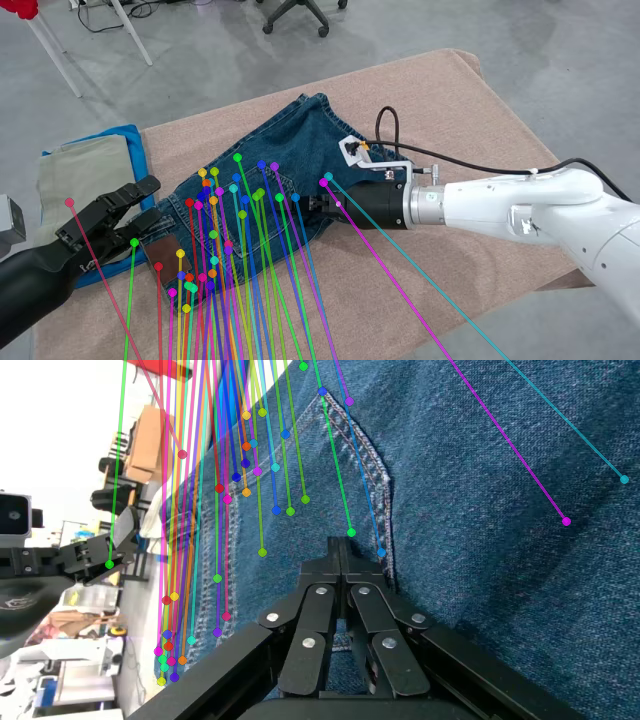}
        \caption{$\mathcal{P}_{\mathrm{view}}$}
        \label{fig:calibration-view}
    \end{subfigure}
    \caption{Correspondences used for calibration: (\subref{fig:calibration-align}) robot renderings and observations, (\subref{fig:calibration-time}) static backgrounds across time, and (\subref{fig:calibration-view}) synchronized camera views.}
    \label{fig:calibration-constraints}
    \vspace{-10pt}
\end{wrapfigure}

We first render the robot's URDF model using recorded
robot states and initial camera parameters, projecting
the known robot geometry into the image space of the
recorded videos.
We then establish pixel correspondences between the
rendered and observed robots, forming
$\mathcal{P}_{\mathrm{align}}$
(\Cref{fig:calibration-align}).
To further constrain the calibration, we incorporate
static-background correspondences across frames,
$\mathcal{P}_{\mathrm{time}}$ (\Cref{fig:calibration-time}), and correspondences
between synchronized views, $\mathcal{P}_{\mathrm{view}}$
(\Cref{fig:calibration-view}).
We extract correspondences using RoMaV2~\citep{edstedt_roma_2025}, with masks
from SAM3~\citep{carion_sam_2026} fine-tuned on robot segmentation data
restricting robot--observation matching to arm regions.

\paragraph{Joint calibration.}
Using the three correspondence sets, we jointly refine
camera parameters $\Theta$, including intrinsics,
extrinsics, and distortion, together with relative
arm mounting offsets $\xi$ where applicable.
For $\mathcal{P}_{\mathrm{align}}$, we minimize the
reprojection error between
$\pi_i(X_i(s,\xi);\Theta,\xi)$ and its matched pixel
$\vu_i$ in the recorded frame.
The matched rendering pixel identifies a point
on the known robot surface, whose 3D position
$X_i(s,\xi)$ is determined by the recorded robot state
$s$ and arm mounting offset $\xi$.
The projection $\pi_i$ uses camera parameters $\Theta$
and, for arm-mounted cameras, also depends on $\xi$.
For $\mathcal{P}_{\mathrm{time}}$ and
$\mathcal{P}_{\mathrm{view}}$, we impose the same
ray-coplanarity constraint: rays $\vr_a,\vr_b$
observing the same scene point must be coplanar
with the baseline between their camera centers
$\vo_a,\vo_b$.
These constraints give the joint objective
\begin{equation}
\scalebox{0.9}{$\displaystyle
(\Theta^*, \xi^*)
= \arg\min_{\Theta,\xi}\;
\lambda_a
\left\langle
\left\|
\vu_i-\pi_i\!\left(X_i(s,\xi);\Theta,\xi\right)
\right\|_2^2
\right\rangle_{\mathcal{P}_{\mathrm{align}}}
+
\lambda_e
\left\langle
\left|
(\vr_a\times\vr_b)^\top(\vo_a-\vo_b)
\right|
\right\rangle_{\mathcal{P}_{\mathrm{view}}
\cup\mathcal{P}_{\mathrm{time}}}.
$}
\label{eq:calibration}
\end{equation}
Here, $\langle\cdot\rangle_{\mathcal{P}}$ denotes an average over correspondences in $\mathcal{P}$,
and $\lambda_a,\lambda_e$ balance the two terms. The optimized parameters
$(\Theta^*,\xi^*)$
are then used to construct robot renderings for action conditioning.

\subsection{Stage I: Supervised Training with Cross-Embodiment Action Representation}
\label{sec:stage1}

\begin{figure}[htbp]
    \centering
    \includegraphics[width=\linewidth]{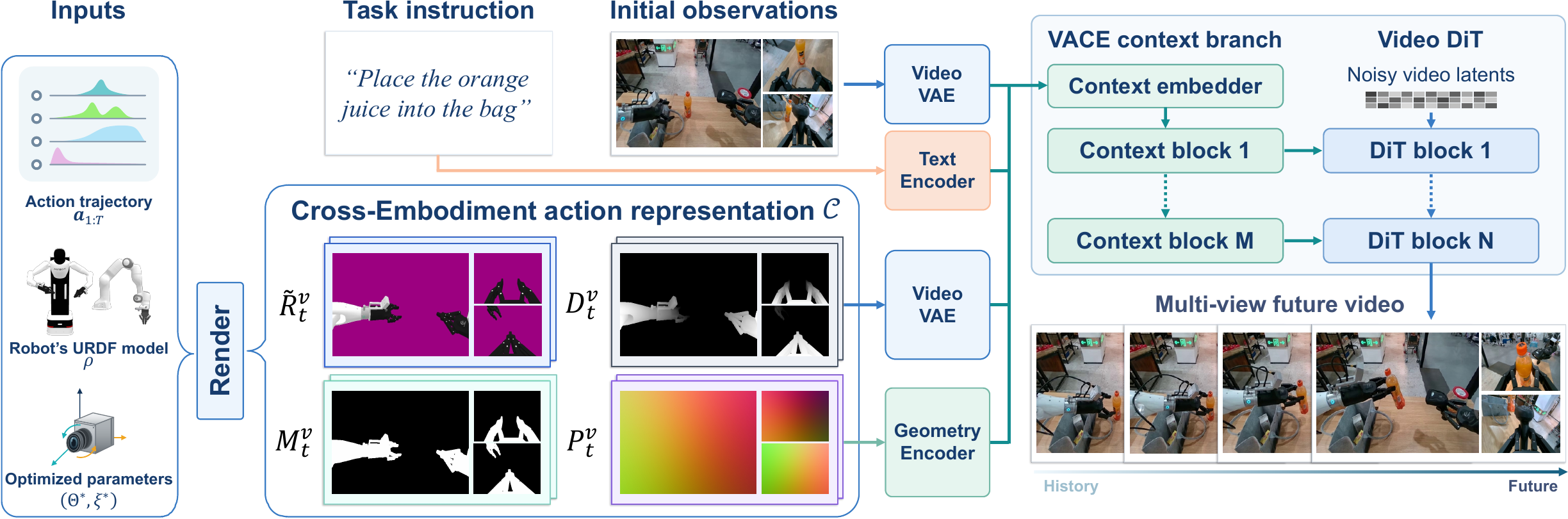}
    \caption{Unified multi-view world model. Robot renderings, gripper openings, and camera rays condition the video DiT through a VACE branch to predict multi-view future videos.}
    \label{fig:method_stage1}
\end{figure}

Stage I learns action-conditioned future prediction
across robot embodiments using a unified image-space action
representation (\cref{fig:method_stage1}).

\paragraph{Cross-Embodiment action representation.}
Actions are defined differently across robot embodiments,
making it difficult to use the same action encoder
across datasets.
We therefore convert these specific actions into image-space
conditions in a common format, allowing actions from
different embodiments to be encoded in the same way.
Specifically, using the robot's URDF model $\rho$
and the optimized parameters
$(\Theta^*,\xi^*)$,
we render the robot according to the action trajectory
$\va_{1:T}$, producing RGB images $R_t^v$,
depth maps $D_t^v$, and amodal masks $M_t^v$
at each time step $t$ and view $v$. 
We additionally encode gripper states and camera motion to provide state cues beyond the robot renderings.
Gripper openings are linearly mapped to background
RGB intensities in $[0,255]$, yielding $\tilde{R}_t^v$,
while camera rays are encoded as dense Pl\"ucker
maps $P_t^v$. The resulting action representation is:
$\mathcal{C}=\{(\tilde{R}_t^v,D_t^v,M_t^v,P_t^v)\}_{t=1:T,\,v=1:V}$.

\paragraph{Action-conditioned training.}
We incorporate the action representation $\mathcal{C}$
into the generator through a
VACE~\citep{jiang_vace_2025}
conditioning branch.
The rendered RGB images and normalized depth maps
are separately encoded by the pretrained video VAE,
while the masks and Pl\"ucker maps are jointly encoded
by a 3D convolutional geometry encoder.
The resulting features are concatenated to form the
input context of the VACE branch, which produces
residuals injected into the corresponding
DiT~\citep{peebles_dit_2023} layers.
For multi-view prediction, video latents from different
views are tiled along the width dimension in a fixed
order, with the action features arranged in the same
layout. Initial multi-view observations are encoded
as reference latent frames.

We train the generator on multiple robot datasets $\mathcal{D}$, using paired future videos
as supervision. Let $z$ denote the ground-truth multi-view
video latent, $z_\tau$ its noisy version at time $\tau$
with sampled noise $\epsilon$, and $u_\tau$ the corresponding
flow-matching~\citep{lipman_flow_matching_2022}
velocity target. Given the condition
$c=(\vx_0^{1:V},\ell,\mathcal{C})$,
we train the generator $f_\theta$ with the objective
\begin{equation}
\mathcal{L}_{\mathrm{SFT}}
=
\mathbb{E}_{(z,c)\sim\mathcal{D},\,\tau,\,\epsilon}
\left[
\left\|
f_\theta(z_\tau,\tau;c)-u_\tau
\right\|_2^2
\right].
\label{eq:sft}
\end{equation}

\subsection{Stage II: Counterfactual Post-Training with Embodied Video Rewards}
\label{sec:method_stage2}

Stage II extends training with counterfactual actions,
exposing the model to a broader range of action and
contact configurations.
We learn a video reward model from human annotations
of generation defects and use its feedback to guide
reinforcement learning~\citep{
zhu_diffusion_rl_survey_2023,black_ddpo_2024}.

\paragraph{Counterfactual action construction.} 
We modify a recorded trajectory by applying an SE(3)
perturbation to its final end-effector pose and
interpolating from the fixed initial pose to the
perturbed endpoint.
Inverse kinematics~\citep{guo_mplib_2021}
converts this trajectory into a counterfactual
action sequence, which we use to construct
the conditioning inputs $\mathcal{C}$
as described in~\cref{sec:stage1}.
The initial observations are fixed to explore
the predicted outcomes of different actions
from the same scene.

\begin{wrapfigure}{r}{0.48\textwidth}
    \centering
    \includegraphics[width=\linewidth]{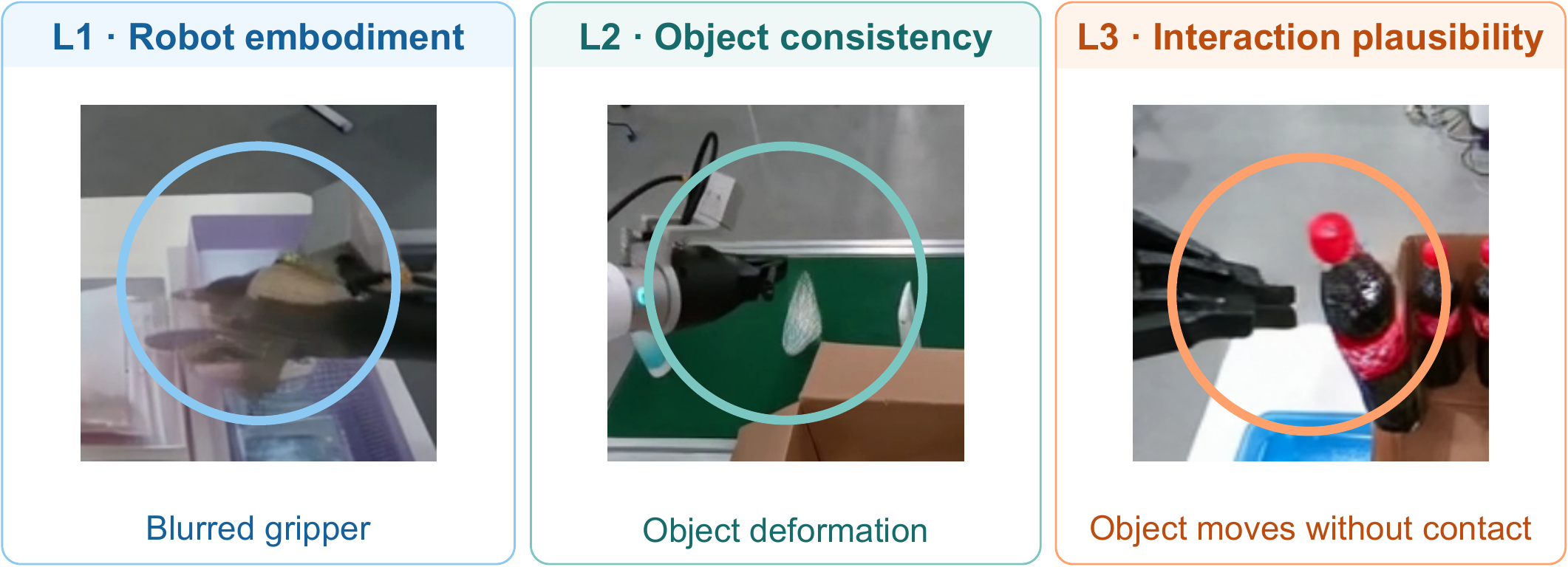}
    \caption{Representative defects across three dimensions: robot embodiment (L1), object consistency (L2), and interaction plausibility (L3).}
    \label{fig:method_reward_model}
    \vspace{-13pt}
\end{wrapfigure}

\paragraph{Embodied video reward learning.}
Although counterfactual actions lack paired ground-truth
future videos, the physical plausibility of their
generated predictions can still be assessed through
observable defects. For example, an object rising with
a gripper after a missed grasp reveals an interaction
error without requiring a reference video.
To capture defects in both the robot and objects
themselves and their interactions, we construct
a human-annotated video defect dataset using
predictions from multiple world models under
recorded and counterfactual actions.
The annotations cover three complementary dimensions:
robot embodiment (L1), object consistency (L2),
and interaction plausibility (L3), as illustrated
in \Cref{fig:method_reward_model}.

Using these annotations, we fine-tune a vision-language
reward model~\citep{qwen_qwen35_omni_2026} $q_\phi$
with cross-entropy loss to predict defects in each
dimension. Given a generated video $\hat{x}$ and a
fixed evaluation prompt, let $p_\phi^k(\hat{x})$
denote the predicted defect probability for
$k\in\{\mathrm{L1},\mathrm{L2},\mathrm{L3}\}$.
We define the continuous reward as
$R^k(\hat{x})=-p_\phi^k(\hat{x})$,
so that lower predicted defect probabilities yield
higher rewards. This converts binary annotations
into continuous feedback for generator post-training.
Reward-model data and annotation details are provided
in \Cref{app:video-judge}.

\paragraph{Reward-guided generator post-training.}
Starting from Stage I, we post-train the generator by augmenting the recorded trajectories with the counterfactual actions constructed above. 
For each resulting action condition, we sample a group of future
videos and score them with the frozen reward model
$q_\phi$. All predictions receive the three defect-based
rewards. Predictions under recorded actions additionally
receive a PSNR reward against their paired ground-truth
futures to preserve fidelity.
Following GDPO~\citep{liu_gdpo_2026}, we normalize
each reward within the sampled group, combine the
resulting advantages using a weighted sum, and apply
batch-wise normalization to obtain $A_i$.
We then optimize the generator with DiffusionNFT~\citep{zheng_diffusionnft_2025}:
\begin{equation}
\mathcal{L}_{\mathrm{NFT}}
=
\mathbb{E}_{i,\tau,\epsilon_i}
\left[
r(A_i)\ell_i^+
+
\bigl(1-r(A_i)\bigr)\ell_i^-
\right].
\end{equation}
Here, $\tau$ and $\epsilon_i$ denote the sampled
noise time and noise, respectively.
The positive and negative denoising objectives
$\ell_i^+$ and $\ell_i^-$ and the clipped-advantage
mapping $r(\cdot)$ follow DiffusionNFT.

\section{Experiments}
\label{sec:experiments}

\subsection{Experimental Setup}
\label{sec:experimental-setup}

\paragraph{Datasets and benchmarks.}
\method is jointly trained on
AgiBotWorld-Beta~\citep{bu_agibot_2025},
DROID~\citep{khazatsky_droid_2024},
RoboMIND~2.0~\citep{hou_robomind_2025}, and
RoboTwin~2.0~\citep{mu_robotwin_2025}.
After filtering, the training data contain
2,232\,h of multi-view trajectories across five
robot arm types in real and simulated environments
(\Cref{fig:data-pipeline}, in appendix).
We evaluate future prediction on held-out episodes
from these datasets.
On AgiBot, we additionally construct 160
counterfactual test conditions across 54 tasks
by perturbing recorded trajectories as described
in ~\cref{sec:method_stage2}.

\paragraph{Baselines and metrics.}
We compare against DreamDojo~\citep{gao_dreamdojo_2026},
GE-Sim~2.0~\citep{qiu_gesim2_2026},
Genie Envisioner~\citep{liao_genie_envisioner_2025},
and EnerVerse-AC~\citep{jiang_enerverse_ac_2025},
using their native conditioning interfaces.
We also compare our Stage-I model
(\emph{Ours (w/o RL)}) with the full Stage-II model
(\emph{Ours (full)}) to assess the effect of
reward-guided post-training.
We measure action following using normalised
dynamic time warping (\ndtw) from
EWMBench~\citep{yue_ewmbench_2025}, and visual
fidelity against ground-truth videos using
PSNR~\citep{hore_psnr_2010},
SSIM~\citep{wang_ssim_2004}, and
LPIPS~\citep{zhang_lpips_2018}.
We additionally report EWMScore-P, which aggregates
15 metrics across six WorldArena quality
dimensions~\citep{shang_worldarena_2026}.
To assess physical plausibility, three independent
annotators inspect counterfactual predictions for
defects in robot embodiment, object consistency,
and interaction plausibility.
We report defect rates based on majority agreement
(Fleiss' $\kappa = 0.7817$ for any-defect labels).
Further evaluation details are provided in
\Cref{app:evaluation-protocol}.

\subsection{Action Following and Interaction Plausibility}
\label{sec:exp-agibot-main}

\begin{table}[ht]
    \caption{Benchmark evaluation on AgiBot under recorded and counterfactual actions. Defect rates are reported in percent (\%); \rankbest{best}, \ranksecond{second-best}, and \rankthird{third-best} scores are highlighted in blue.}
    \label{tab:modified-main}
    \centering
    \small
    \setlength{\tabcolsep}{4pt}
    \resizebox{\linewidth}{!}{\begin{tabular}{lccccc cccc}
        \toprule
        & \multicolumn{5}{c}{Recorded actions} & \multicolumn{4}{c}{Counterfactual actions} \\
        \cmidrule(lr){2-6}\cmidrule(lr){7-10}
        Model & EWMScore-P $\uparrow$ & PSNR $\uparrow$ & SSIM $\uparrow$ & LPIPS $\downarrow$ & \ndtw $\uparrow$ & Embodiment $\downarrow$ & Object $\downarrow$ & Interaction $\downarrow$ & \ndtw $\uparrow$ \\
        \midrule
        EnerVerse-AC & 68.61 & 19.17 & 0.863 & 0.158 & 0.8058 & 1.25 & 49.38 & 48.12 & 0.6828 \\
        Genie Envisioner & 70.37 & \rankthird{20.72} & \rankthird{0.890} & \rankthird{0.112} & \rankthird{0.8124} & 0.00 & 43.75 & 54.38 & 0.7379\\
        DreamDojo & 70.01 & 19.27 & 0.812 & 0.163 & 0.6578 & 0.00 & \rankthird{12.50} & \rankthird{16.25} & 0.6407\\
        GE-Sim 2.0 & \rankthird{72.36} & 19.91 & 0.874 & 0.130 & 0.8063 & 0.00 & \ranksecond{3.75} & \ranksecond{7.50} & \rankthird{0.7817}\\
        \midrule
        Ours (w/o RL) & \rankbest{72.84} & \ranksecond{22.66} & \ranksecond{0.928} & \ranksecond{0.099} & \ranksecond{0.8711} & 0.00 & 31.88 & 48.12 & \ranksecond{0.8735}\\
        Ours (full) & \ranksecond{72.51} & \rankbest{23.06} & \rankbest{0.936} & \rankbest{0.094} & \rankbest{0.8772} & 0.00 & \rankbest{3.12} & \rankbest{6.25} & \rankbest{0.8831}\\
        \bottomrule
    \end{tabular}}
\end{table}

\begin{figure}[htbp]
    \centering
    \includegraphics[width=\linewidth]{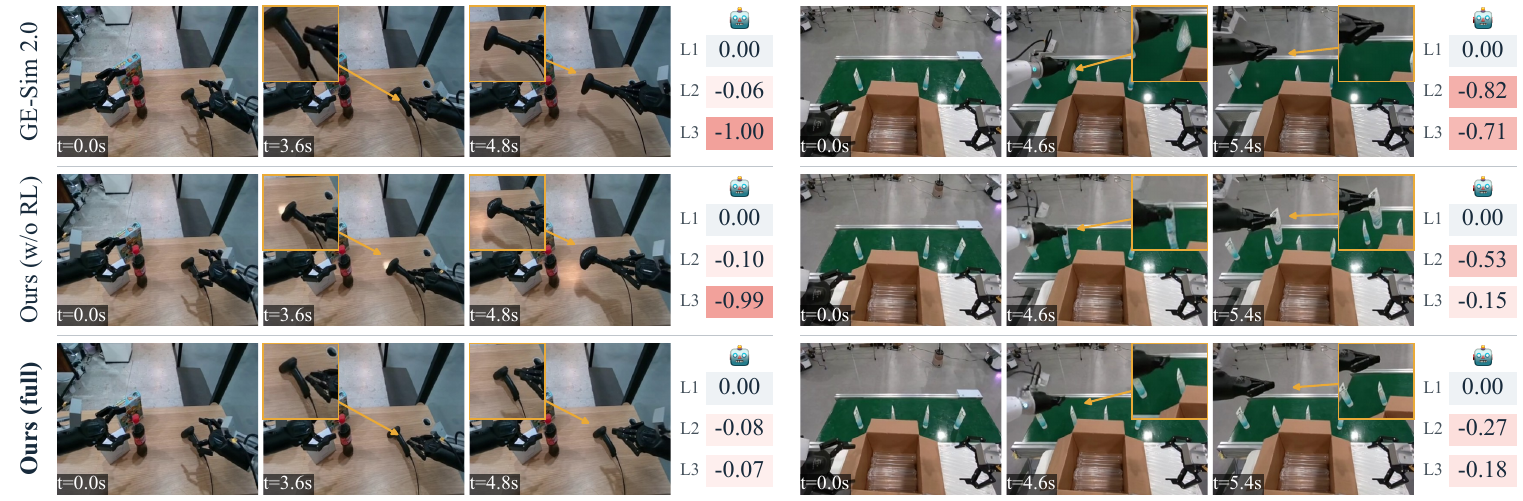}
    \caption{Qualitative comparison of counterfactual action predictions. Compared with GE-Sim 2.0 and our model without RL, the full model avoids unsupported scanner motion and object disappearance or duplication in these examples. Yellow annotations highlight relevant regions. The reward model scores L1--L3 assess embodiment, object, and interaction quality, respectively.}
    \label{fig:main-physical-plausibility}
\end{figure}

\begin{wraptable}{r}{0.47\linewidth}
    \vspace{-13pt}
    \centering
    \caption{Action-conditioned generation results of the world model track of the AgiBot World Challenge 2026. Best results are in \textbf{bold}; second-best results are \underline{underlined}.}
    \label{tab:agibot-challenge2026}

    \small
    \setlength{\tabcolsep}{2.5pt}
    \renewcommand{\arraystretch}{0.78}
    \newcommand{\challengemedal}[2]{\begingroup
        \setlength{\unitlength}{1pt}\begin{picture}(10,9)(0,1)
            \put(4,4){\color[HTML]{4779B5}\rotatebox[origin=bl]{27}{\rule{1.5pt}{5pt}}}
            \put(4,4){\color[HTML]{C94C4C}\rotatebox[origin=bl]{-27}{\rule{1.5pt}{5pt}}}
            \put(5,3){\color[HTML]{#1}\circle*{6}}
            \put(5,3){\color{black}\makebox(0,0){\fontsize{4}{4}\selectfont\textbf{#2}}}
        \end{picture}\endgroup\,}
    \resizebox{\linewidth}{!}{\begin{tabular}{lccc|c}
        \toprule
        Team & Visual Quality $\uparrow$ & Scene Consistency $\uparrow$ & Action Following $\uparrow$ & EWMScore $\uparrow$ \\
        \midrule
        \challengemedal{B87333}{3}Loop & \underline{0.6207} & \underline{0.9024} & 0.9492 & 0.8241 \\
        \challengemedal{B8BEC8}{2}PAIWorld & 0.6161 & \textbf{0.9041} & \underline{0.9531} & \underline{0.8245} \\
        \midrule
        \challengemedal{E2B642}{1}Ours & \textbf{0.6246} & 0.8974 & \textbf{0.9651} & \textbf{0.829} \\
        \bottomrule
    \end{tabular}}
    \vspace{-10pt}
\end{wraptable}

\method accurately follows the supplied actions
while maintaining visual fidelity.
As shown in \Cref{tab:modified-main}, the full model achieves the
highest \ndtw among evaluated methods under both
recorded and counterfactual actions, with scores
of 0.8772 and 0.8831, respectively.
It also achieves the best PSNR, SSIM, and LPIPS
against ground-truth videos.
Our model ranks first in the world model track
of the AgiBot World Challenge 2026, achieving the
highest overall score, action following, and visual
quality among participating teams (\Cref{tab:agibot-challenge2026}).

\begin{figure}[htbp]
    \centering
    \includegraphics[width=\linewidth]{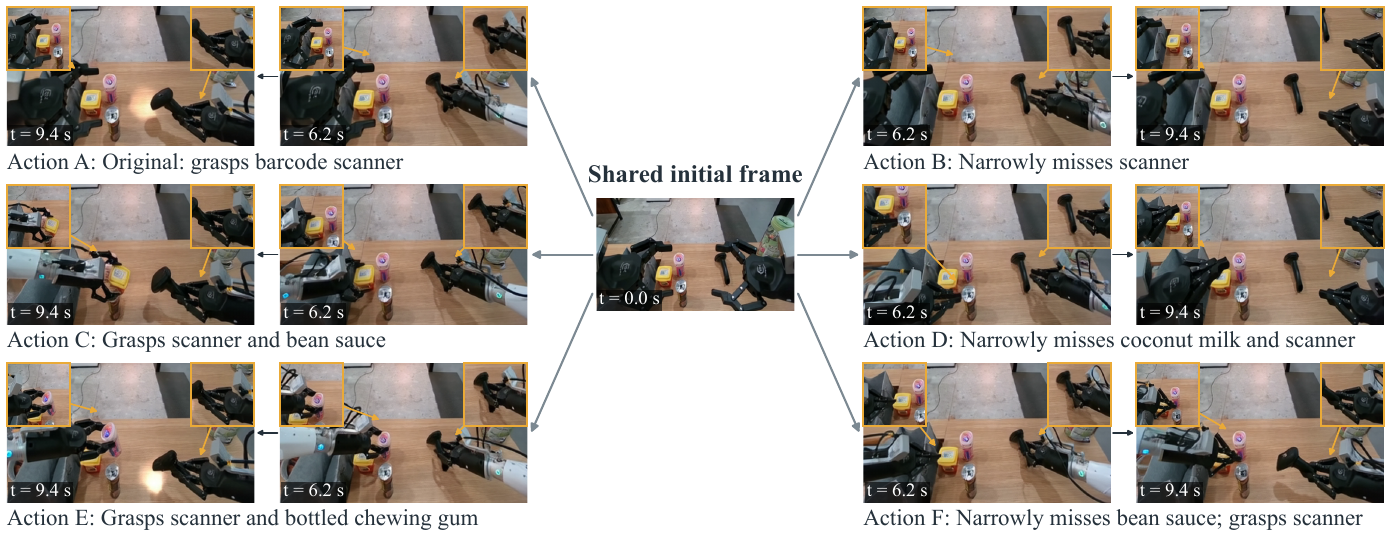}
    \caption{Predictions from the same initial scene under multiple action conditions.}
    \label{fig:multi-conditions}
\end{figure}

We further evaluate whether objects respond
coherently to the supplied actions.
Compared with Stage I, reward-guided post-training
reduces object defects from 31.88\% to 3.12\%
and interaction defects from 48.12\% to 6.25\% (\Cref{tab:modified-main}). 
Meanwhile, \ndtw remains comparable, indicating
that interaction quality improves without
compromising action following.
\cref{fig:main-physical-plausibility} illustrates the reduced defects: the full model leaves an object on the table after a missed
grasp and avoids the object disappearance or
duplication observed in the comparison examples.
\cref{fig:multi-conditions} further examines interaction outcomes
when the initial scene is fixed and the actions
vary, covering different grasp targets and
single- and dual-arm manipulation.
Grasped objects move with the grippers, while
ungrasped objects remain on the table, including
cases where only one arm successfully grasps.
These examples illustrate physically plausible
object responses under different actions.

\subsection{Prediction across Embodiments and Environments}
\label{sec:exp-other-datasets}

\begin{figure}[htbp]
    \centering
    \includegraphics[width=\linewidth]{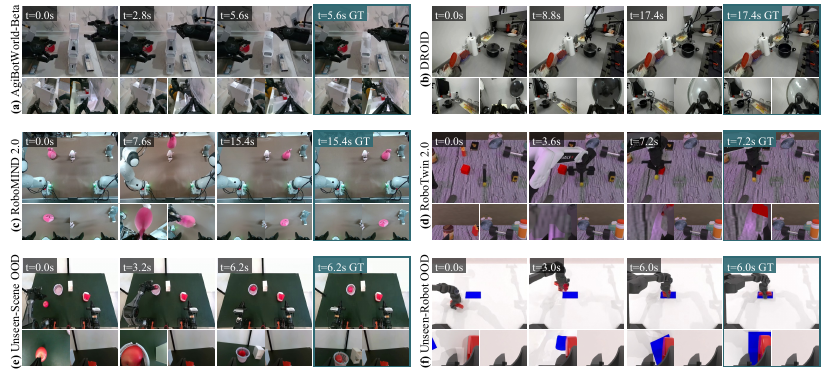}
    \caption{Prediction across embodiments and environments. A shared checkpoint generates rollouts on 4 datasets and generalises without additional fine-tuning to (e) unseen real-world scenes and (f) an unseen robot embodiment (WidowX250).}
    \label{fig:visual}
    \vspace{-12pt}
\end{figure}

\begin{wraptable}{r}{0.54\linewidth}
    \centering
    \vspace{-12pt}
    \caption{Multi-view recorded-action fidelity on DROID, RoboMIND 2.0, and RoboTwin 2.0.}
    \label{tab:normal-main}
    \vspace{-4pt}
    \small
    \resizebox{\linewidth}{!}{\begin{tabular}{llccc}
        \toprule
        Dataset & Model & PSNR $\uparrow$ & SSIM $\uparrow$ & LPIPS $\downarrow$ \\
        \midrule
        \multirow{2}{*}{DROID}
        & Ctrl-World & \underline{22.00} & \underline{0.772} & \underline{0.162} \\
        & Ours (full) & \textbf{22.95} & \textbf{0.848} & \textbf{0.073} \\
        \midrule
        \multirow{1}{*}{RoboMIND 2.0}
        & Ours (full) & \textbf{23.84} & \textbf{0.826} & \textbf{0.126} \\
        \midrule
        \multirow{1}{*}{RoboTwin 2.0}
        & Ours (full) & \textbf{27.35} & \textbf{0.834} & \textbf{0.155} \\
        \bottomrule
    \end{tabular}}
    \vspace{-10pt}
\end{wraptable}

Our model supports prediction across different
robot embodiments and environments, as assessed
on held-out episodes from DROID, RoboMIND 2.0,
and RoboTwin 2.0 (\Cref{tab:normal-main}).
On DROID, it outperforms CtrlWorld across all
three fidelity metrics, improving PSNR from
22.00 to 22.95 and reducing LPIPS from 0.162
to 0.073.
\Cref{fig:visual}(a--d) shows representative
predictions across all four training datasets,
covering real and simulated robot manipulation.

We further examine transfer to an unseen robot
embodiment and unseen real-world scenes.
For embodiment transfer, we introduce WidowX250,
which is excluded from training, into RoboTwin 2.0.
For scene transfer, we evaluate self-collected
Piper recordings in unseen real-world scenes.
Without additional fine-tuning, the model produces
the predictions shown in \Cref{fig:visual}(e,f), providing
qualitative evidence of transfer in both settings.
Extended video sequences are provided in \Cref{app:ood-rollouts}.

\subsection{Policy Outcome Evaluation}
\label{sec:policy-evaluation}

\begin{wrapfigure}{r}{0.45\linewidth}
    \vspace{-10pt}
    \centering
    \includegraphics[width=\linewidth]{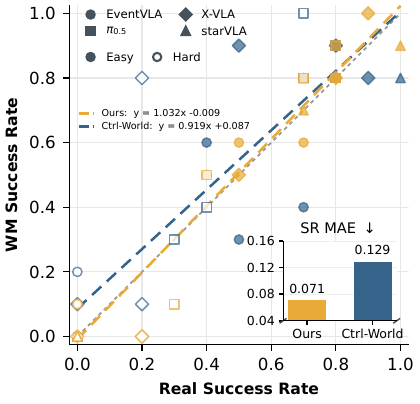}
    \caption{Policy outcome alignment against ground-truth simulation across 24 evaluation conditions (inset: success-rate MAE).}
    \label{fig:policy-eval-alignment}
    \vspace{-20pt}
\end{wrapfigure}

Following world-model-based policy evaluation, we examine whether generated rollouts faithfully reproduce the success rates and rankings of robotic policies without spurious completions. Across three bimanual RoboTwin~2.0 tasks~\citep{mu_robotwin_2025} under easy and hard settings, each world model predicts visual outcomes conditioned on 240 action trajectories supplied by four VLA policies, comparing human-annotated video outcomes against ground-truth simulation labels (\Cref{app:policy-eval-details}).

As visualized in \Cref{fig:policy-eval-alignment}, our model yields a linear fit that virtually matches the ideal diagonal ($y = 1.032x - 0.009$), whereas baseline predictions exhibit a substantial upward shift ($y = 0.919x + 0.087$).
This divergence is most evident in the low-success regime, where baseline rollouts consistently float above the diagonal by hallucinating completions on failed grasps.
In contrast, our predictions cluster tightly along the diagonal across both easy and hard settings, cutting success-rate MAE from 12.92 to 7.08 percentage points (pp) as shown in the inset.
Across the 24 task--policy--setting combinations, our model achieves strong rank correlation (Spearman $\rho = 0.937$) and near-zero aggregate bias (+0.42 pp).

\subsection{Effect of Geometric Calibration}
\label{sec:exp-calibration}

\begin{table}[ht]
    \caption{Ablation on geometric calibration across training and inference stages on AgiBot and DROID. Sync error (Synchronous Position Error) measures the 2D Euclidean distance (in pixels) between predicted and ground-truth end-effector positions across temporally aligned frames.}
    \label{tab:ablations}
    \centering
    \small
    \resizebox{\linewidth}{!}{\begin{tabular}{lllccc|cc}
        \toprule
        Dataset & Train calib. & Infer calib. & EWMScore-P $\uparrow$ & PSNR $\uparrow$ & SSIM $\uparrow$ & \ndtw $\uparrow$ & Sync error (px) $\downarrow$ \\
        \midrule
        AgiBot & Dataset-provided & Dataset-provided & 72.11 & 21.06 & 0.898 & 0.8159 & 7.73 \\
        AgiBot & Dataset-provided & Refined & \underline{72.38} & \underline{21.38} & \underline{0.907} & \underline{0.8243} & \underline{5.77} \\
        AgiBot & Refined & Refined & \textbf{72.84} & \textbf{22.66} & \textbf{0.928} & \textbf{0.8711} & \textbf{4.60} \\
        \midrule
        DROID & Dataset-provided & Dataset-provided & 71.03 & 20.77 & 0.809 & 0.7718 & 11.78 \\
        DROID & Dataset-provided & Refined & \underline{71.51} & \underline{21.36} & \underline{0.817} & \underline{0.8569} & \underline{10.21} \\
        DROID & Refined & Refined & \textbf{72.29} & \textbf{22.66} & \textbf{0.841} & \textbf{0.8919} & \textbf{8.06} \\
        \bottomrule
    \end{tabular}}
\end{table}

\begin{figure}[htbp]
    \centering
    \captionsetup[subfigure]{font=calibrationsmall,skip=3pt}
    \begin{subfigure}[t]{0.245\linewidth}
        \vspace{0pt}
        \centering
        \includegraphics[width=0.49\linewidth]{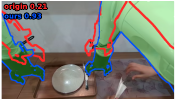}\hfill
        \includegraphics[width=0.49\linewidth]{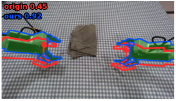}\par
        \vspace{2pt}
        \includegraphics[width=0.49\linewidth]{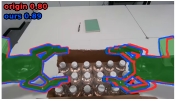}\hfill
        \includegraphics[width=0.49\linewidth]{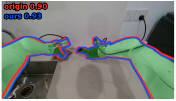}
        \caption{AgiBot examples}
        \label{fig:calibration-agibot-examples}
    \end{subfigure}\hfill
    \begin{subfigure}[t]{0.245\linewidth}
        \vspace{0pt}
        \centering
        \includegraphics[width=0.49\linewidth]{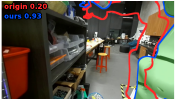}\hfill
        \includegraphics[width=0.49\linewidth]{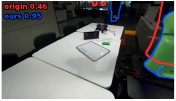}\par
        \vspace{2pt}
        \includegraphics[width=0.49\linewidth]{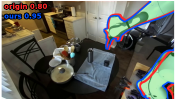}\hfill
        \includegraphics[width=0.49\linewidth]{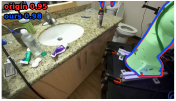}
        \caption{DROID examples}
        \label{fig:calibration-droid-examples}
    \end{subfigure}\hfill
    \begin{subfigure}[t]{0.16\linewidth}
        \vspace{0pt}
        \centering
        \includegraphics[width=\linewidth]{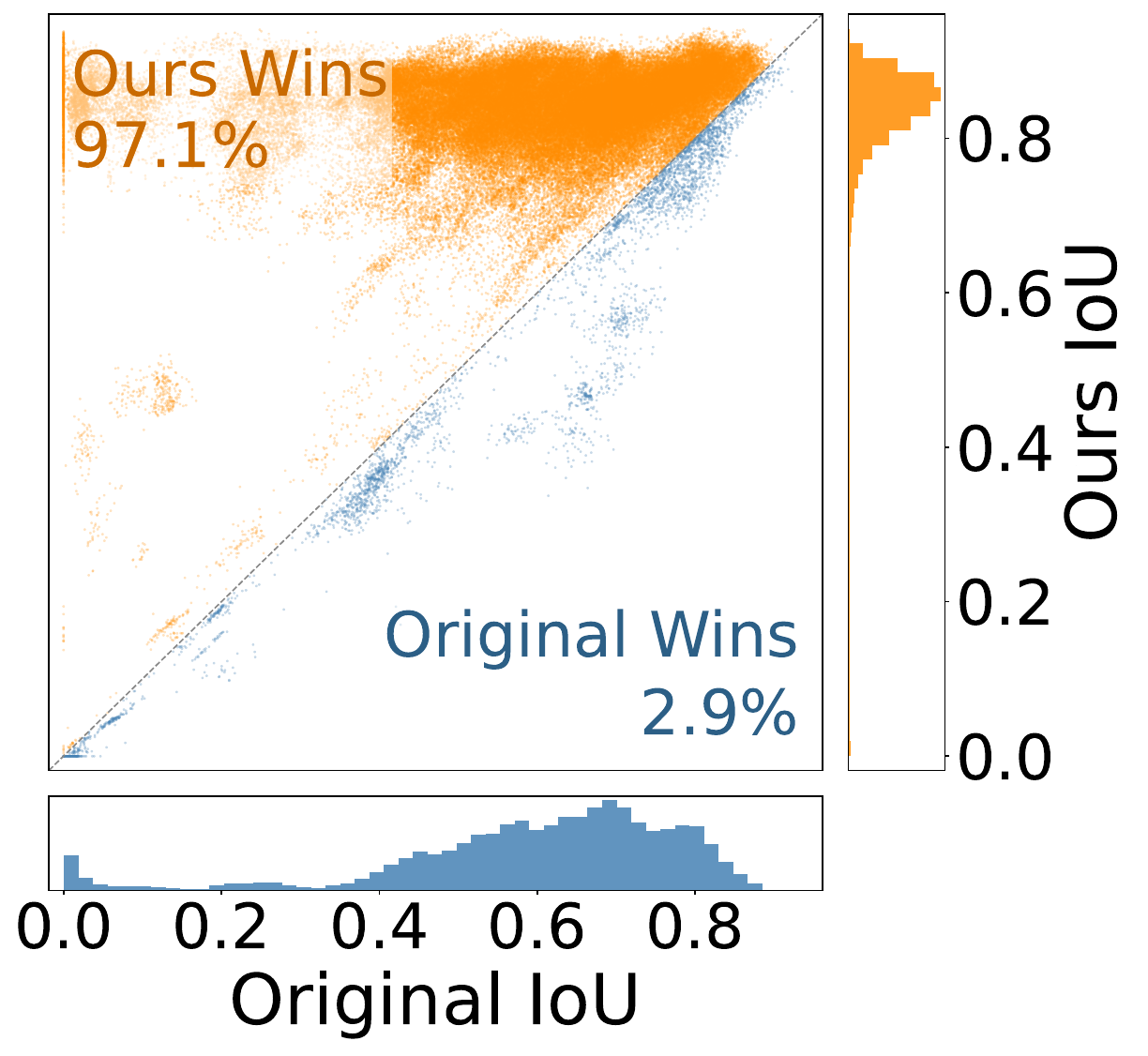}
        \caption{AgiBot: Original vs.\ Ours}
        \label{fig:calibration-agibot-iou}
    \end{subfigure}\hfill
    \begin{subfigure}[t]{0.16\linewidth}
        \vspace{0pt}
        \centering
        \includegraphics[width=\linewidth]{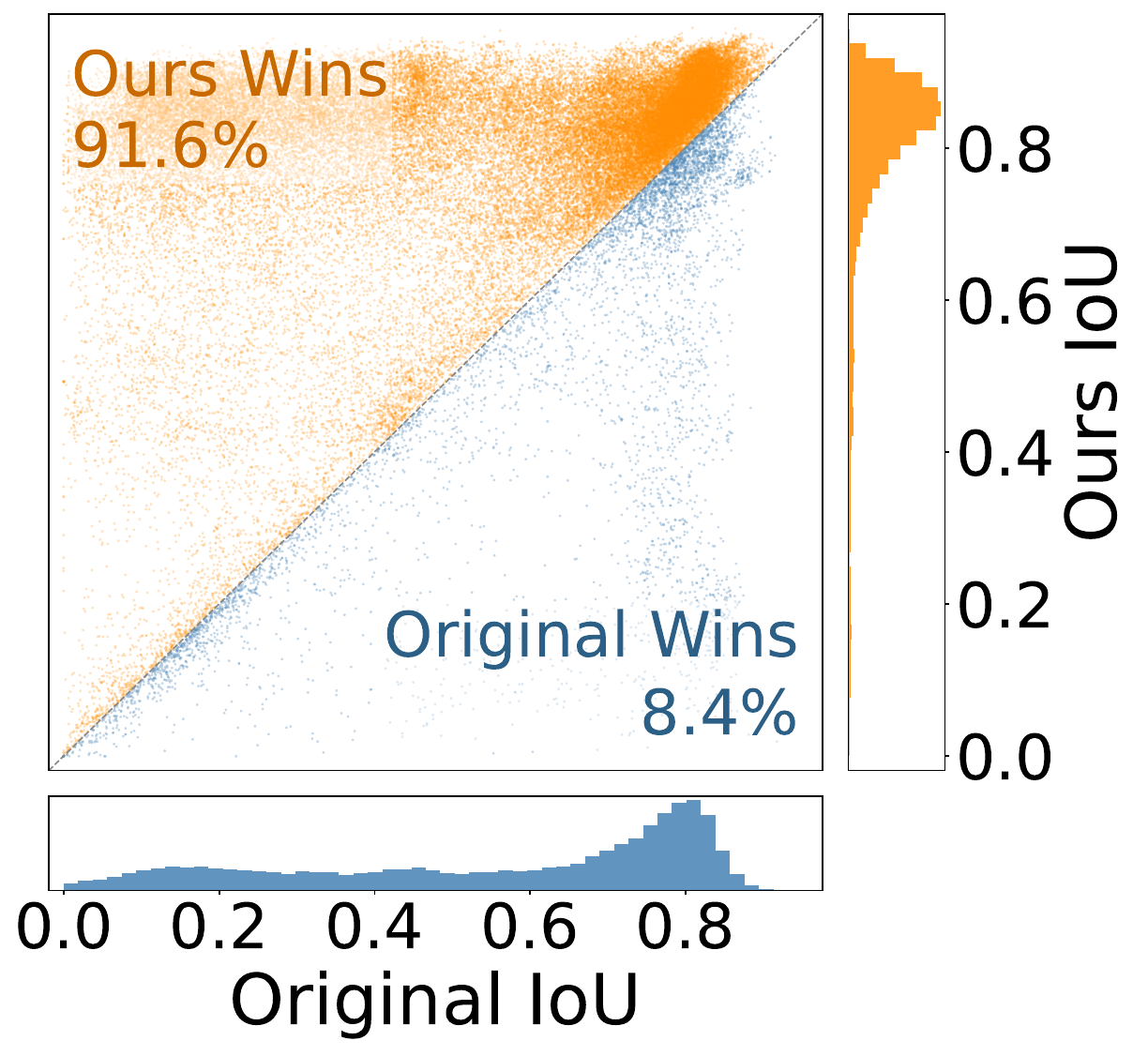}
        \caption{DROID: Original vs.\ Ours}
        \label{fig:calibration-droid-iou}
    \end{subfigure}\hfill
    \begin{subfigure}[t]{0.16\linewidth}
        \vspace{0pt}
        \centering
        \includegraphics[width=\linewidth]{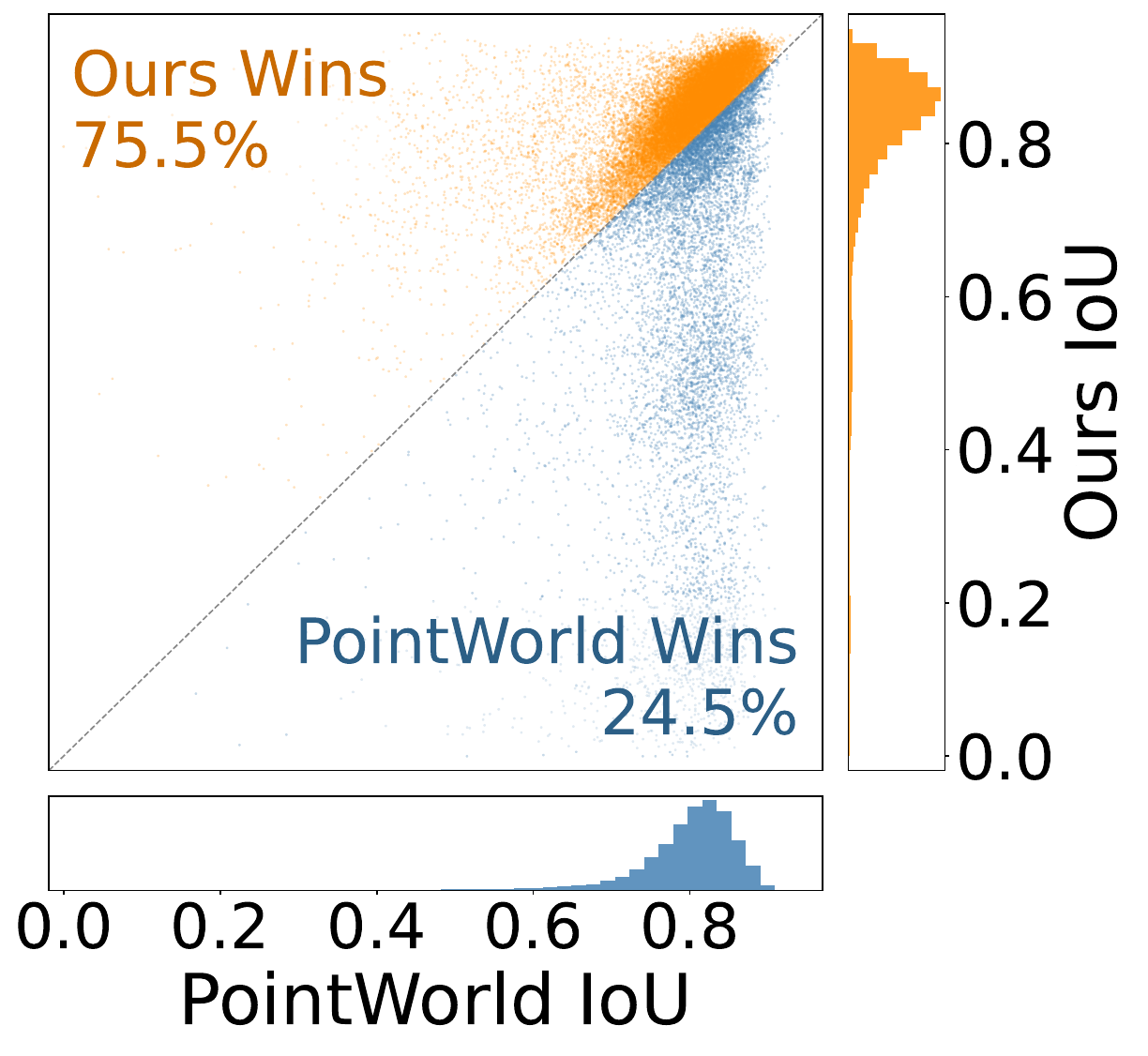}
        \caption{DROID: PointWorld vs.\ Ours}
        \label{fig:calibration-droid-pw-iou}
    \end{subfigure}
    \caption{Geometric calibration results. (\subref{fig:calibration-agibot-examples},\subref{fig:calibration-droid-examples}) SAM3 masks (green) overlaid with Original (red) and Ours (blue) contours. (\subref{fig:calibration-agibot-iou}--\subref{fig:calibration-droid-pw-iou}) IoU comparisons against Original and PointWorld, using AgiBot's head view and DROID's two external view mean. Points above the diagonal favor Ours.}
    \label{fig:calibration-results}
\end{figure}

We evaluate geometric calibration using area-weighted IoU between robot renderings and SAM3 masks. Relative to dataset-provided calibration, our method improves mean IoU by 0.228 on 130,182 AgiBot episodes and 0.212 on 63,061 DROID episodes, with improvements on 97.1\% and 91.6\% of episodes, respectively (\Cref{fig:calibration-results}).
On DROID, we additionally compare against PointWorld~\citep{huang_pointworld_2026}. Despite using RGB only, our method outperforms PointWorld on 75.5\% of 38,356 episodes (\Cref{fig:calibration-droid-pw-iou}); PointWorld relies on FoundationStereo~\citep{wen_foundationstereo_2025} depth and therefore requires stereo inputs, which is unavailable in our other datasets.

\Cref{tab:ablations} further examines how improved alignment affects video prediction. Applying refined calibration only at inference improves all reported metrics on both datasets, indicating that the generator benefits from more accurate action conditions even without retraining. Using refined calibration during training as well
yields further gains: PSNR improves by 1.28\,dB
on AgiBot and 1.30\,dB on DROID, with a consistent
reduction in synchronous position error across
both datasets. These results support the value of spatial alignment for both inference-time conditioning and learning
from paired actions and videos.

\subsection{Reward Model Evaluation}
\label{sec:exp-reward}

\begin{wraptable}{r}{0.50\linewidth}
    \centering
    \vspace{-12pt}
    \caption{Defect discrimination performance of the embodied video reward model on the held-out validation benchmark. Accuracy and macro-F1 are reported in percent (\%).}
    \label{tab:vlm-validation}
    \vspace{-4pt}
    \small
    \resizebox{\linewidth}{!}{\begin{tabular}{lccc}
        \toprule
        Dimension & Accuracy $\uparrow$ & Macro-F1 $\uparrow$ & MAE $\downarrow$ \\
        \midrule
        L1: Embodiment & 96.69 & 69.37 & 0.041 \\
        L2: Object & 79.24 & 79.00 & 0.243 \\
        L3: Interaction & 83.08 & 74.27 & 0.198 \\
        \midrule
        Average & 86.33 & 74.21 & 0.160 \\
        \bottomrule
    \end{tabular}}
    \vspace{-8pt}
\end{wraptable}

We assess whether the embodied video reward model
identifies defects in agreement with human judgments
on 4,893 held-out manipulation clips
(\Cref{tab:vlm-validation}).
Data splits and metric definitions are provided
in \Cref{app:video-judge}.
Across the three dimensions, the model achieves
an average accuracy of 86.33\% and macro-F1
of 74.21\%.
The per-dimension results on L1--L3 further
support the model's ability to identify robot,
object, and interaction defects, providing
complementary feedback for Stage-II post-training.
The effect of this feedback on generation quality
is evaluated independently by human annotators
in \Cref{sec:exp-agibot-main}.

\section{Conclusion and Limitations}
\label{sec:conclusion}

We present \method, a multi-view, cross-embodiment
robot world model for action-faithful and physically
plausible video prediction.
Offline geometric calibration improves action following
and prediction fidelity by aligning rendered action
conditions with recorded videos, while counterfactual
post-training improves robot--object interaction
plausibility through broader interaction coverage
and feedback from a learned embodied video reward model.
Experiments further show that the resulting predictions
enable more accurate estimation of policy outcomes,
supporting the use of world models for evaluating
robot actions before execution.
We also release refined calibration results
for 153,666 episodes across three datasets and
a video defect annotation corpus covering 44.9K videos from multiple world models.
\paragraph{Limitations.}
The action representation, generator, and VLM reward model all operate in image space; under occlusion or limited views, the framework may generate or reward visually plausible but physically incorrect interactions.

\clearpage
\bibliography{references_world_models,references_calibration}
\bibliographystyle{iclr2027_conference}

\clearpage
\appendix
\section*{SUPPLEMENTARY MATERIAL}
\section{Model, Data, and Condition Construction}
\label{app:model-data-condition}

The following details specify the implementation and condition interface used
for the reported experiments.

Throughout the paper, \emph{Ours (w/o RL)} denotes the supervised Stage-I
checkpoint, while \emph{Ours (full)} denotes the Stage-II checkpoint obtained
by counterfactual post-training from the same Stage-I model.

\subsection{Implementation and Training}
\label{app:implementation}

The video generator, initialized from
Wan2.1-VACE-14B, processes 101-frame clips with three synchronized views at
$240\times320$ resolution per view and a frame-aligned shared visual action
representation.
It conditions on the first RGB frame from each view and the task instruction
to predict the remaining 100 frames. DROID uses two external cameras and one
gripper camera; the other three-view layouts use a primary camera and two
wrist cameras. Video latents are tiled along the width dimension in a fixed
order, with action features arranged in the same layout.
Stage I trains the DiT and VACE LoRA branches together with the 3D
convolutional geometry encoder for masks and \plucker maps. Stage II keeps the
base generator, the geometry encoder, and the reward model frozen and updates
only the DiT and VACE LoRA branches. The model has approximately 0.7B
trainable parameters. The LoRA rank is 128 and the Stage-II learning rate is
$5\times10^{-6}$. The frozen Qwen3.5-9B reward model supplies
L1, L2, and L3 rewards for both types of conditions; PSNR is added only for
recorded actions with paired ground-truth futures. GDPO normalizes each reward
within a group, combines the normalized rewards
with a weighted sum, and applies batch-wise normalization. The resulting
advantages weight the positive and negative denoising objectives of
DiffusionNFT. The normalized L1, L2, and L3 reward channels are equally
weighted with a weight of 1 each, while the recorded-action PSNR channel has
a weight of 0.5.

Stage I jointly samples AgiBotWorld-Beta, DROID, RoboMIND~2.0, and
RoboTwin~2.0, comprising the 2,232\,h of filtered trajectories described in
\Cref{sec:experimental-setup}. The shared action representation accommodates
their different embodiments and camera layouts. Stage II reuses recorded conditions together with their
counterfactual action edits; paired-future fidelity is applied only to samples
with an observed future. No dataset-specific fine-tuning is
used for the cross-embodiment results.

\subsection{Condition Construction}
\label{app:condition-construction}

\subsubsection{Implementation Details Offline Geometric Calibration}

For real-world data, offline geometric calibration is applied before condition
rendering. It jointly refines camera intrinsics, extrinsics, distortion, and,
where applicable, arm mounting offsets using the
correspondence losses in \Cref{sec:camera-refinement}. Robot--image
alignment uses render--observation correspondences, while temporal and
cross-view correspondences impose ray coplanarity. Calibration quality is
evaluated by area-weighted IoU between rendered robot masks and SAM3 masks
over 130,182 AgiBot episodes and 63,061 DROID episodes, as reported in
\Cref{sec:exp-calibration,fig:calibration-results}. The calibration ablation
separately considers refinement at inference only and at both training and
inference.

We group episodes within each real-world dataset by available collection
metadata, such as collector, camera, and robot identifiers. Within each group,
we calibrate a randomly selected episode and evaluate the resulting parameters
on other episodes using randomly sampled frames. We progressively calibrate
additional episodes with poor alignment for each group. AgiBot and RoboMIND use a fixed
initialization, whereas DROID uses 32 initializations to accommodate varying
camera extrinsics. Arm mounting offsets are optimized for RoboMIND and for the
self-collected Piper recordings used in unseen-scene OOD evaluation. For
RoboTwin and the WidowX250 embodiment transferred into it, we use ground-truth
geometry from the simulator.

Evaluation episodes undergo calibration-quality checks and may receive additional calibration when alignment is poor. 
Calibration of evaluation episodes does not use ground-truth frames from the prediction interval.
The calibrated geometry is frozen and used to render the action conditions for video prediction.

\begin{figure}[htbp]
    \centering
    \includegraphics[width=\linewidth]{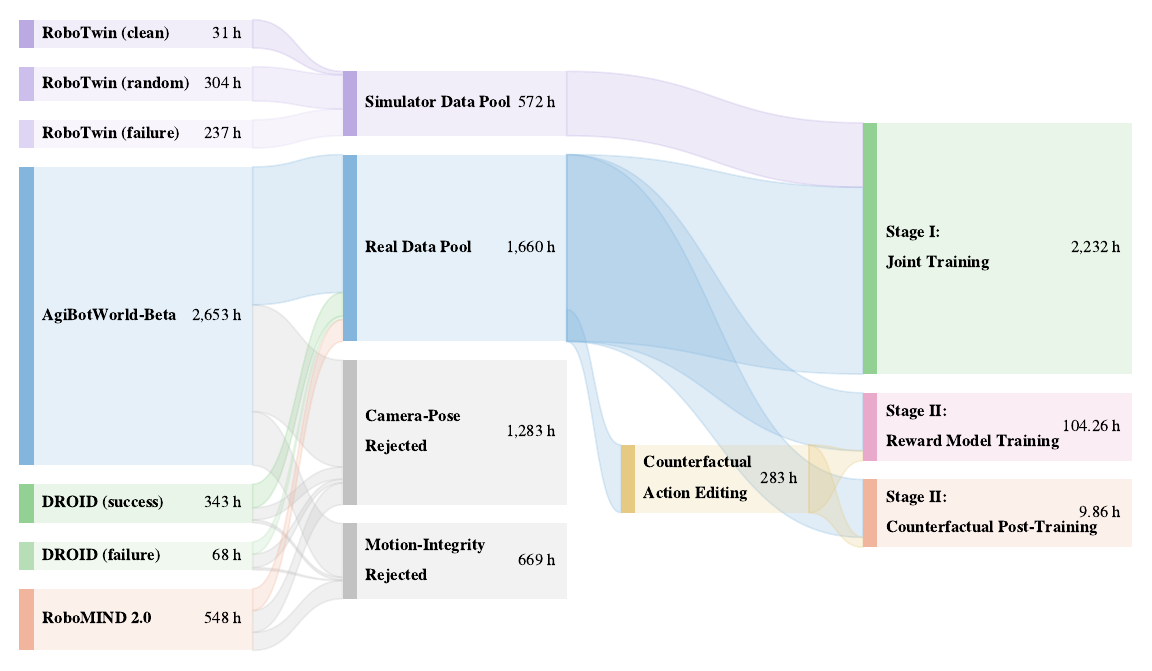}
    \caption{Training-data curation pipeline from raw demonstration sources through geometric calibration and motion filtering to the two training stages; numbers denote cumulative hours.}
    \label{fig:data-pipeline}
\end{figure}

\paragraph{Released calibration.}
We release per-episode calibration parameters for the filtered real-data pool shown in \Cref{fig:data-pipeline}, covering 153,666 episodes and 1,660.31\,h across AgiBotWorld-Beta (1,225.60\,h), DROID (239.73\,h), and RoboMIND~2.0 (194.98\,h). Each record contains camera intrinsics, extrinsics, and distortion parameters. DROID and RoboMIND~2.0 records additionally contain mounting offsets and calibration-quality metrics.

\subsubsection{Rendered Channels}

For every frame and view, the preprocessing pipeline stores robot RGB, depth,
mask, camera metadata, and robot state. Robot RGB and normalized depth are
encoded by the video VAE, while the robot mask and \plucker map are processed
by the 3D convolutional geometry encoder. Link-level segmentation and simulator metadata
support preprocessing and evaluation but are not model inputs.

\subsubsection{Gripper Channel Convention}

Gripper openings are linearly mapped to background RGB intensities in
$[0,255]$, as in Stage I. The left and right gripper values occupy the red
and green channels, respectively. These continuous state cues are included
in the rendered RGB condition before VAE encoding.

\subsubsection{Counterfactual Trajectory Construction}

An SE(3) perturbation is applied to the final end-effector pose of a
demonstrated trajectory. We interpolate from the fixed initial pose to this
perturbed endpoint and use inverse kinematics to obtain the modified joint
sequence. Initial observations and the task instruction remain unchanged.
Counterfactual edits retain the original episode's fixed calibration parameters.
Robot renderings and trajectory-dependent wrist-camera geometry are
recomputed from the modified sequence to construct the same shared action
representation used in Stage I. Only kinematically feasible conditions are
retained.

\section{Reward Model, Annotation, and Metrics}
\label{app:reward-annotation-metrics}

\subsection{Embodied Video Reward Model and Human Annotation}
\label{app:video-judge}

The annotation form contains three dimensions: L1 Embodiment, L2 Object, and
L3 Interaction. L1 records duplication, disappearance, and structural
or material defects; L2 records object appearance, disappearance, deformation,
material, and texture defects; and L3 records contact--motion mismatch and
physically implausible interaction. For reward-model training, all three
dimensions use the binary labels \emph{none} and \emph{defective}, derived
from a single annotation per generated video. For downstream human evaluation,
defect rates are computed from video-level majority labels provided by three
independent annotators, as specified in \Cref{app:evaluation-protocol}.

The reward model predicts the three top-level labels in one structured
response. A single response supplies three label distributions at their
respective token positions. For each dimension $k$, the reward is the negative
defect-label probability, $R^k(\hat{x})=-p_\phi^k(\hat{x})$, as defined in
Stage II. The three channels are normalized independently by GDPO.
We fine-tune all parameters of Qwen3.5-9B with an autoregressive
cross-entropy loss weighted per sample and per dimension. The weights balance
task--label groups within each source variant and then balance total weights
across variants for each dimension. The corpus combines paired ground-truth videos,
used as defect-free references, with human-annotated predictions from several
world models under recorded or kinematically valid counterfactual action
conditions. Videos are sampled at 5\,FPS, resized to a height of 240 pixels,
tiled in their existing multi-view layout.
The model receives only the video being assessed and a fixed evaluation prompt at
scoring time, without an additional paired source future or rendered action condition.

\paragraph{Data composition and split.}
The reward-model corpus combines ground-truth videos with annotated predictions
under recorded and counterfactual actions. It covers AgiBot~\citep{bu_agibot_2025},
DROID~\citep{khazatsky_droid_2024}, RoboMIND~\citep{hou_robomind_2025},
and RoboTwin~\citep{mu_robotwin_2025}, and additionally includes predictions generated from single-view
Bridge dataset~\citep{walke_bridgedata_2023}. Bridge is used only for
reward-model training and validation, not for training the multi-view video
generator. Predictions come from our model and seven external models:
DreamDojo~\citep{gao_dreamdojo_2026},
Ctrl-World~\citep{guo_ctrl-world_2025},
EnerVerse-AC~\citep{jiang_enerverse_ac_2025},
Genie Envisioner~\citep{liao_genie_envisioner_2025},
GE-Sim 2.0~\citep{qiu_gesim2_2026},
Cosmos Predict 2.5~\citep{nvidia_world_2025}, and IRASim~\citep{zhu_irasim_2025}.
The training and validation sets contain 31{,}600 and 4{,}893 distinct video
paths, respectively (\Cref{tab:reward-data}), and are split by source-episode
identifier.

\paragraph{Released annotations.}
The annotation corpus covers 44.9K original videos and contains 30.4K defect labels across the L1, L2, and L3 dimensions. After filtering, 36,493 distinct video paths are used for reward-model training and validation, including 31,600 training videos and 4,893 validation videos.

\begin{table}[htbp]
    \centering
    \caption{Reward-model data composition, counted by distinct video paths.}
    \label{tab:reward-data}
    \small
    \begin{tabular}{lrr}
        \toprule
        Video source & Training & Validation \\
        \midrule
        Ground-truth videos & 3{,}680 & 667 \\
        Recorded-action predictions & 16{,}067 & 2{,}090 \\
        Counterfactual-action predictions & 11{,}853 & 2{,}136 \\
        \midrule
        Total & 31{,}600 & 4{,}893 \\
        \bottomrule
    \end{tabular}
\end{table}

Validation uses the 4{,}893-video set reported in
\Cref{tab:vlm-validation} with the fixed joint prompt. The reward model is
frozen during Stage-II generator post-training. Downstream defect rates in
\Cref{tab:modified-main} are assessed independently by human annotators.

\paragraph{Probability-based MAE.}
For video $i$ and dimension $k$, $p_i^k$ denotes the defect probability
predicted by the reward model, and $y_i^k\in\{0,1\}$ is the corresponding binary
annotation, where 1 indicates a defect. The MAE in
\Cref{tab:vlm-validation} is
\begin{equation}
    \mathrm{MAE}_k=\frac{1}{N}\sum_{i=1}^{N}\left|p_i^k-y_i^k\right|,
    \qquad N=4{,}893.
\end{equation}

\subsection{nDTW Action-Following Metric}
\label{app:action-metric}

We use \ndtw following EWMBench~\citep{yue_ewmbench_2025} to evaluate action
following. This metric is used only for evaluation and does not enter Stage-II
post-training. It measures gripper-trajectory agreement; Embodiment, Object, and Interaction defects are evaluated
separately.

\section{Evaluation and Baseline Protocol}
\label{app:evaluation-baseline}

\subsection{Evaluation and Human Annotation Details}
\label{app:evaluation-protocol}

\paragraph{Counterfactual benchmark.}
Counterfactual evaluation uses the same 160 conditions across 54 AgiBot tasks
for all methods. We evaluate outputs from DreamDojo, GE-Sim~2.0, Genie
Envisioner, EnerVerse-AC, Ours (w/o RL), and Ours (full).
\Cref{tab:modified-main} reports the benchmark results under recorded and
counterfactual actions.

\paragraph{Recorded-action evaluation sets.}
The AgiBot recorded-action evaluation in \Cref{tab:modified-main} uses the
same 800 episodes for all methods and nDTW. The recorded-action evaluations in
\Cref{tab:normal-main} use 50 episodes each from DROID, RoboMIND~2.0, and
RoboTwin~2.0. All evaluation episodes are held out from training.

\begin{table}[t]
    \caption{WorldArena quality on recorded actions. EWMScore-P aggregates the six displayed quality dimensions. Best, second-best, and third-best scores are highlighted.}
    \label{tab:worldarena-quality}
    \centering
    \small
    \textbf{WorldArena quality (recorded actions)}\\[2pt]
    \resizebox{\linewidth}{!}{\begin{tabular}{lcccccc|c}
        \toprule
        Model & Visual $\uparrow$ & Motion $\uparrow$ & Content $\uparrow$ & Physics $\uparrow$ & 3D $\uparrow$ & Control $\uparrow$ & EWMScore-P $\uparrow$ \\
        \midrule
        EnerVerse-AC & 64.72 & 38.14 & 72.12 & 76.56 & 94.99 & 80.56 & 68.61 \\
        Genie Envisioner & 70.22 & 36.47 & 73.25 & 78.38 & \ranksecond{97.46} & 82.00 & 70.37 \\
        DreamDojo & \rankbest{71.41} & 36.44 & 74.13 & 75.19 & \rankthird{96.71} & 80.21 & 70.01 \\
        GE-Sim 2.0 & \rankthird{70.62} & \rankbest{41.01} & \rankthird{74.83} & \rankbest{81.72} & \rankbest{97.87} & \rankthird{83.41} & \rankthird{72.36} \\
        \midrule
        Ours (w/o RL) & \ranksecond{70.73} & \ranksecond{40.73} & \ranksecond{75.61} & \ranksecond{81.40} & 96.55 & \ranksecond{87.79} & \rankbest{72.84} \\
        Ours (full) & 69.30 & \rankthird{40.24} & \rankbest{75.62} & \rankthird{81.36} & 96.66 & \rankbest{88.09} & \ranksecond{72.51} \\
        \bottomrule
    \end{tabular}}
\end{table}

\paragraph{Baseline protocol.}
Each method receives the same initial observation, task instruction, and
counterfactual action semantics through its native conditioning interface.
All methods are evaluated on the same condition set using the same metrics;
our model receives the rendered robot conditions and camera-ray encodings
described in \Cref{app:condition-construction}.

\paragraph{Human evaluation.}
Three annotators independently assess each of the 960 counterfactual videos
(160 conditions for each of the six evaluated models) for Embodiment, Object,
and Interaction defects. Model identities are hidden from annotators, and the
video presentation order is randomized independently for each annotator.
The annotation records contain one assessment from each annotator for every
video; all 960 videos therefore have complete triplicate annotations. We do
not introduce a separate repeat-pass or adjudication label. As a data-integrity
check, we found no missing or duplicate annotator--video records. Annotations are
aggregated at the video level: a video is labeled defective in a dimension if
at least two annotators identify a defect in that dimension. A video may have
defects in multiple dimensions, so these fractions may overlap.

Table~\ref{tab:human-agreement} reports observed agreement and Fleiss
$\kappa$ for any defect and for the three annotation dimensions.

Table~\ref{tab:human-ci} reports majority-vote defect rates with Wilson 95\%
confidence intervals. Because all models use the same conditions, pairwise
comparisons use two-sided exact McNemar tests on majority labels; these tests
are descriptive and are not corrected for multiple comparisons.

\begin{table}[t]
    \caption{Inter-annotator agreement on the 960 counterfactual videos used in Table~\ref{tab:modified-main}. Observed agreement is the mean fraction of agreeing annotator pairs per video.}
    \label{tab:human-agreement}
    \centering
    \small
    \setlength{\tabcolsep}{7pt}
    \begin{tabular}{lcc}
        \toprule
        Dimension & Observed agreement (\%) & Fleiss $\kappa$ \\
        \midrule
        Any defect & 89.72 & 0.7817 \\
        L1: Embodiment & 99.65 & 0.4427 \\
        L2: Object & 89.31 & 0.7135 \\
        L3: Interaction & 87.15 & 0.7031 \\
        \bottomrule
    \end{tabular}
\end{table}

\begin{table}[t]
    \caption{Majority-vote counterfactual defect rates with Wilson 95\% confidence intervals. Each rate uses the same 160 conditions as Table~\ref{tab:modified-main}.}
    \label{tab:human-ci}
    \centering
    \scriptsize
    \setlength{\tabcolsep}{3.5pt}
    \begin{tabular}{lccc}
        \toprule
        Model & L1: Embodiment (\%) & L2: Object (\%) & L3: Interaction (\%) \\
        \midrule
        EnerVerse-AC & 1.25 [0.34, 4.44] & 49.38 [41.73, 57.05] & 48.12 [40.52, 55.82] \\
        Genie Envisioner & 0.00 [0.00, 2.34] & 43.75 [36.30, 51.49] & 54.38 [46.65, 61.90] \\
        DreamDojo & 0.00 [0.00, 2.34] & 12.50 [8.24, 18.52] & 16.25 [11.34, 22.75] \\
        GE-Sim 2.0 & 0.00 [0.00, 2.34] & 3.75 [1.73, 7.94] & 7.50 [4.34, 12.65] \\
        Ours (w/o RL) & 0.00 [0.00, 2.34] & 31.88 [25.15, 39.45] & 48.12 [40.52, 55.82] \\
        Ours (full) & 0.00 [0.00, 2.34] & 3.12 [1.34, 7.11] & 6.25 [3.43, 11.12] \\
        \bottomrule
    \end{tabular}
\end{table}

\subsection{Policy Evaluation Details}
\label{app:policy-eval-details}

\paragraph{Tasks and environment settings.}

We evaluate policy outcomes on three representative dual-arm manipulation tasks from RoboTwin~2.0~\citep{mu_robotwin_2025} using the Aloha-AgileX bimanual embodiment: \texttt{beat\_block\_hammer}, \texttt{handover\_block}, and \texttt{place\_dual\_shoes}. Together, these tasks cover diverse bimanual manipulation behaviors, including tool use, handover, and coordinated object placement.

We construct 60 frozen evaluation scenarios (10 distinct random seeds for each of the three tasks under two visual environment settings). The two settings comprise \emph{Easy} (\texttt{clean}), which retains canonical asset textures, and \emph{Hard} (\texttt{randomized}), where tabletop and object surface textures are randomly assigned unseen patterns. Each case defines fixed initial scene seeds, object poses, and language instructions, and was empirically verified to be solvable via expert teleoperation.

\paragraph{Evaluated VLA policies.}
We benchmark four open-source vision-language-action models: EventVLA~\citep{yang_eventvla_2026}, $\pi_{0.5}$~\citep{black_pi05_2025}, X-VLA~\citep{zheng_xvla_2026}, and starVLA~\citep{ye_starvla_alpha_2026}, using their official 50-task co-training checkpoints released by RoboTwin~2.0~\citep{mu_robotwin_2025}. Each policy executes in the simulator across the 60 frozen cases, yielding a total of 240 evaluation episodes ($4\text{ policies} \times 60\text{ cases}$).

\paragraph{Action conditioning and outcome annotation.}
Each world model (Ctrl-World and Ours) predicts multi-view future video rollouts conditioned on the initial observation and the complete action trajectory executed by a given policy.
Human annotators independently evaluate each generated video rollout to determine whether task success conditions defined by RoboTwin were fulfilled (e.g., successful block placement, non-dropped handover, target contact). Annotators are fully blinded to the generating model identity and policy checkpoint, with the video display sequence randomized independently for each annotator. Final success labels are aggregated via majority vote.

\paragraph{Metric definitions and evaluation rationale.}
To comprehensively benchmark world models as reliable offline policy evaluators, we report metrics spanning group-level calibration, ranking consistency, and episode-level classification:
\begin{itemize}[leftmargin=1.5em,itemsep=2pt,topsep=2pt]
    \item \textbf{Success-Rate Mean Absolute Error (MAE)}: Computed across the $K=24$ distinct task--policy--setting combinations as $\text{MAE} = \frac{1}{K}\sum_{k=1}^K |\hat{s}_k - s_k^*|$, where $\hat{s}_k$ and $s_k^*$ denote the world model rollout success rate and ground-truth simulation success rate (each over 10 episodes), respectively. Reported in percentage points ($\text{pp}$). It quantifies the absolute calibration gap between generative evaluation and physics simulation.
    \item \textbf{Signed Bias}: Defined as $\text{Bias} = \frac{1}{K}\sum_{k=1}^K (\hat{s}_k - s_k^*)$ in percentage points. It measures systematic directional distortion: a positive bias indicates ``optimistic hallucination'' where failed manipulations are rendered as spurious completions, whereas a value near zero signifies unbiased outcome reflection.
    \item \textbf{Pearson Correlation ($r$) and Spearman Rank Correlation ($\rho$)}: Pearson $r$ evaluates the linear alignment between predicted and simulated success rates, while Spearman $\rho$ measures rank-order agreement across the 24 task--policy--setting combinations.
    \item \textbf{Episode Accuracy (Acc) and F1 Score}: Computed over all $N=240$ individual episodes treating outcome prediction as binary classification. $\text{Acc} = \frac{\text{TP} + \text{TN}}{N}$ and $\text{F1} = \frac{2\text{TP}}{2\text{TP} + \text{FP} + \text{FN}}$ evaluate trajectory-level discrimination, with F1 balancing performance under class imbalance (40\% ground-truth successes vs. 60\% failures).
\end{itemize}

\begin{table}[ht]
\centering
\caption{Policy outcome evaluation on RoboTwin~2.0. Bold denotes best performance.}
\label{tab:policy-eval}
\small
\setlength{\tabcolsep}{6pt}
\resizebox{\linewidth}{!}{\begin{tabular}{lccccccc}
\toprule
Model & MAE (pp) $\downarrow$ & Bias (pp) & Acc. $\uparrow$ & F1 $\uparrow$ & Pearson $r$ $\uparrow$ & Spearman $\rho$ $\uparrow$ \\
\midrule
Ctrl-World  & 12.92 & +5.42 & 0.838 & 0.810 & 0.860 & 0.850 \\
Ours        & \textbf{7.08} & \textbf{+0.42} & \textbf{0.904} & \textbf{0.881} & \textbf{0.964} & \textbf{0.937} \\
\bottomrule
\end{tabular}}
\vspace{-10pt}
\end{table}

\paragraph{Detailed outcome confusion matrix.}
Across all 240 policy episodes, the ground-truth RoboTwin executions contain 96 successes ($40.0\%$) and 144 failures ($60.0\%$).
\Cref{tab:policy-confusion} details the confusion matrix and episode-level classification metrics for both models.
Ctrl-World produces 26 false positives (predicting success when the policy execution actually failed), resulting in an outcome precision of only $76.15\%$ and an inflated positive bias of $+5.42$ percentage points. In contrast, our model reduces false positives to 12, achieving $87.63\%$ precision, $88.54\%$ recall, and an overall accuracy of $90.42\%$, confirming that physically coherent video prediction significantly suppresses hallucinated task completions.

\begin{table}[h]
\centering
\caption{Episode-level outcome classification breakdown across 240 policy rollouts (96 ground-truth successes, 144 ground-truth failures).}
\label{tab:policy-confusion}
\small
\setlength{\tabcolsep}{6pt}
\begin{tabular}{lcccccccc}
\toprule
Model & TP & FP & FN & TN & Precision (\%) & Recall (\%) & Accuracy (\%) & F1 Score \\
\midrule
Ctrl-World & 83 & 26 & 13 & 118 & 76.15 & 86.46 & 83.75 & 0.810 \\
Ours       & \textbf{85} & \textbf{12} & \textbf{11} & \textbf{132} & \textbf{87.63} & \textbf{88.54} & \textbf{90.42} & \textbf{0.881} \\
\bottomrule
\end{tabular}
\end{table}

\section{Additional Out-of-Distribution Rollouts}
\label{app:ood-rollouts}

\begin{figure}[htbp]
    \centering
    \captionsetup[sub]{justification=centering}
    \begin{subfigure}[t]{0.48\linewidth}
        \centering
        \includegraphics[width=\linewidth]{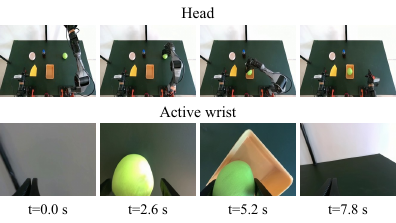}
        \caption{}
    \end{subfigure}\hfill
    \begin{subfigure}[t]{0.48\linewidth}
        \centering
        \includegraphics[width=\linewidth]{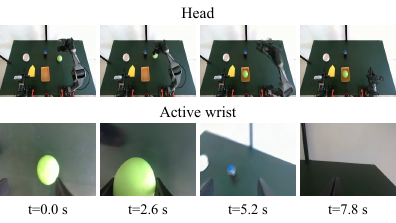}
        \caption{}
    \end{subfigure}\par\vspace{2ex}
    \begin{subfigure}[t]{0.48\linewidth}
        \centering
        \includegraphics[width=\linewidth]{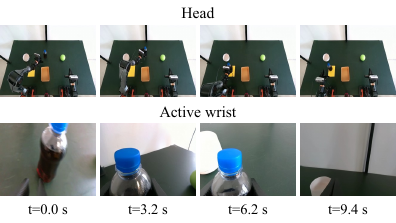}
        \caption{}
    \end{subfigure}\hfill
    \begin{subfigure}[t]{0.48\linewidth}
        \centering
        \includegraphics[width=\linewidth]{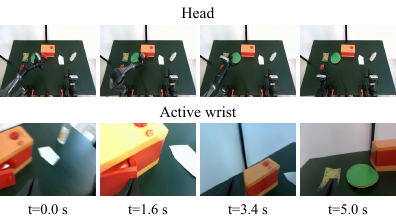}
        \caption{}
    \end{subfigure}\par\vspace{2ex}
    \begin{subfigure}[t]{0.48\linewidth}
        \centering
        \includegraphics[width=\linewidth]{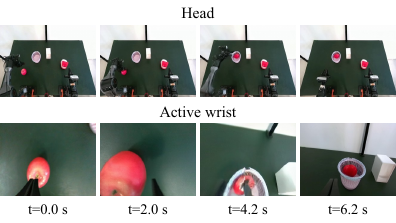}
        \caption{}
    \end{subfigure}\hfill
    \begin{subfigure}[t]{0.48\linewidth}
        \centering
        \includegraphics[width=\linewidth]{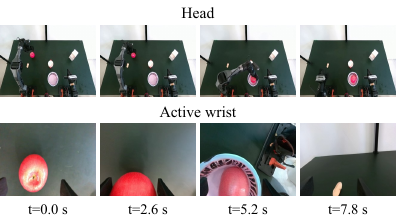}
        \caption{}
    \end{subfigure}
    \caption{\textbf{Self-collected Piper recordings.} Six predicted rollouts, tagged (a)--(f), with synchronized head (top) and active-wrist (bottom) views for each example. Columns progress forward in time, with elapsed video time in seconds below.}
    \label{fig:ood-huairou-all}
\end{figure}

\begin{figure}[t]
    \centering
    \captionsetup[sub]{justification=centering}
    \begin{subfigure}[t]{0.48\linewidth}
        \centering
        \includegraphics[width=\linewidth]{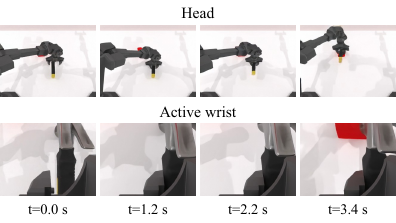}
        \caption{}
    \end{subfigure}\hfill
    \begin{subfigure}[t]{0.48\linewidth}
        \centering
        \includegraphics[width=\linewidth]{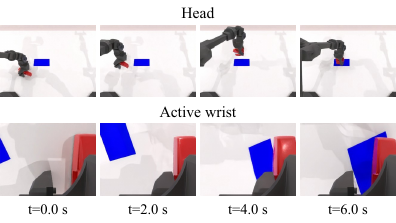}
        \caption{}
    \end{subfigure}\par\vspace{2ex}
    \begin{subfigure}[t]{0.48\linewidth}
        \centering
        \includegraphics[width=\linewidth]{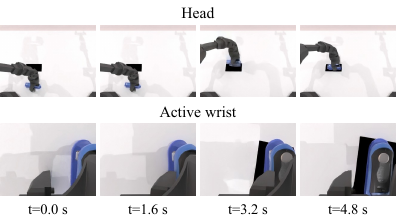}
        \caption{}
    \end{subfigure}\hfill
    \begin{subfigure}[t]{0.48\linewidth}
        \centering
        \includegraphics[width=\linewidth]{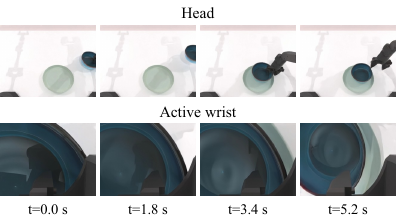}
        \caption{}
    \end{subfigure}\par\vspace{2ex}
    \begin{subfigure}[t]{0.48\linewidth}
        \centering
        \includegraphics[width=\linewidth]{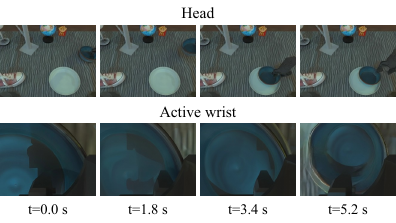}
        \caption{}
    \end{subfigure}\hfill
    \begin{subfigure}[t]{0.48\linewidth}
        \centering
        \includegraphics[width=\linewidth]{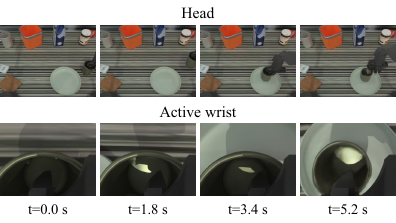}
        \caption{}
    \end{subfigure}
    \caption{\textbf{Unseen-Robot OOD: WidowX250.} Six predicted rollouts, tagged (a)--(f), covering hammering (a), stapler transport (b, c), and container placement in clean (d) and randomized (e, f) scenes. Each example pairs synchronized head (top) and active-wrist (bottom) frames, with columns progressing forward in time.}
    \label{fig:ood-widowx-all}
\end{figure}

\Cref{fig:ood-huairou-all,fig:ood-widowx-all} present all twelve qualitative clips supplied for the two OOD settings: six self-collected Piper recordings and six WidowX250 sequences. Each example shows four sampled time points, including the first and last video frames. Head and active-wrist views are synchronized within each column; $t$ denotes elapsed time in seconds from the start of each prediction video, computed using its frame rate. The inactive wrist view is omitted for compactness. These examples supplement the representative OOD examples in the main evaluation.

\clearpage

\section{Additional Cross-Embodiment Rollouts}
\label{app:cross-embodiment-rollouts}

\begin{figure}[htbp]
    \centering
    \captionsetup[sub]{justification=centering}
    \begin{subfigure}[t]{0.48\linewidth}
        \centering
        \includegraphics[width=\linewidth]{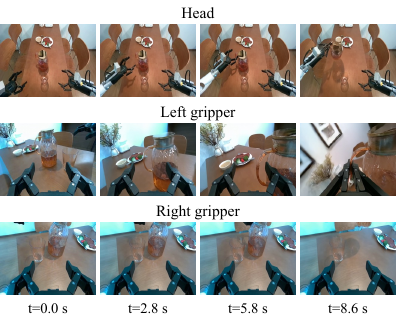}
        \caption{}
    \end{subfigure}\hfill
    \begin{subfigure}[t]{0.48\linewidth}
        \centering
        \includegraphics[width=\linewidth]{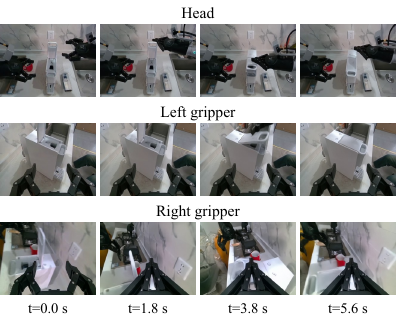}
        \caption{}
    \end{subfigure}\par\vspace{2ex}
    \begin{subfigure}[t]{0.48\linewidth}
        \centering
        \includegraphics[width=\linewidth]{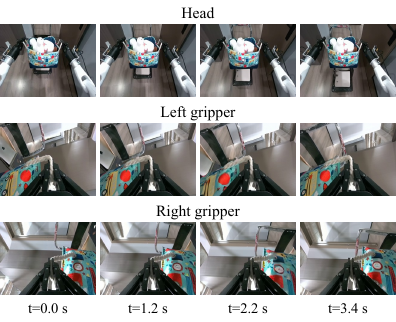}
        \caption{}
    \end{subfigure}\hfill
    \begin{subfigure}[t]{0.48\linewidth}
        \centering
        \includegraphics[width=\linewidth]{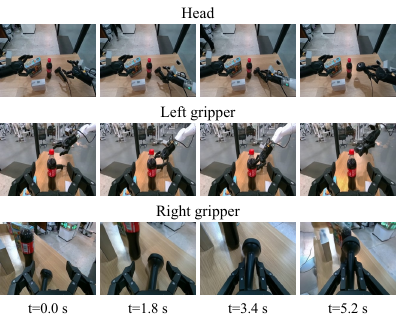}
        \caption{}
    \end{subfigure}\par\vspace{2ex}
    \begin{subfigure}[t]{0.48\linewidth}
        \centering
        \includegraphics[width=\linewidth]{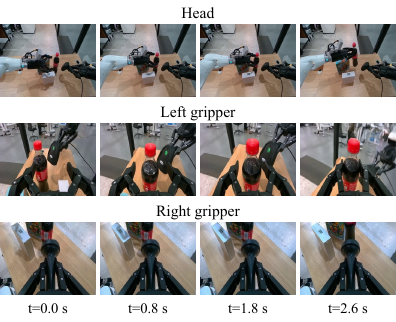}
        \caption{}
    \end{subfigure}\hfill
    \begin{subfigure}[t]{0.48\linewidth}
        \centering
        \includegraphics[width=\linewidth]{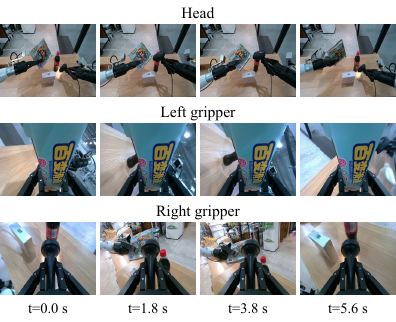}
        \caption{}
    \end{subfigure}
    \caption{\textbf{Cross-embodiment rollouts: AgiBotWorld-Beta.} Six predicted rollouts, tagged (a)--(f), each pairing synchronized head (top), left-gripper (middle), and right-gripper (bottom) views, covering beverage pouring (a, b), object carrying (c), and checkout scenes (d)--(f). Columns progress forward in time, with elapsed video time in seconds below.}
    \label{fig:ce-agibot-all}
\end{figure}

\begin{figure}[t]
    \centering
    \captionsetup[sub]{justification=centering}
    \begin{subfigure}[t]{0.48\linewidth}
        \centering
        \includegraphics[width=\linewidth]{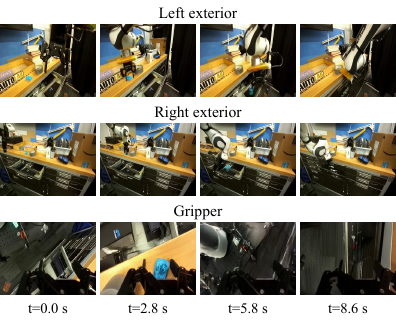}
        \caption{}
    \end{subfigure}\hfill
    \begin{subfigure}[t]{0.48\linewidth}
        \centering
        \includegraphics[width=\linewidth]{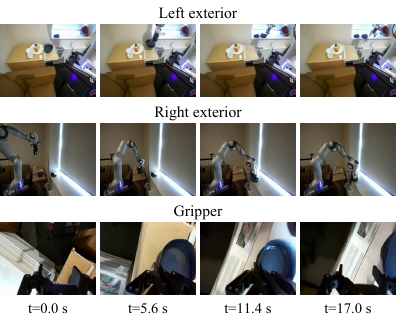}
        \caption{}
    \end{subfigure}\par\vspace{2ex}
    \begin{subfigure}[t]{0.48\linewidth}
        \centering
        \includegraphics[width=\linewidth]{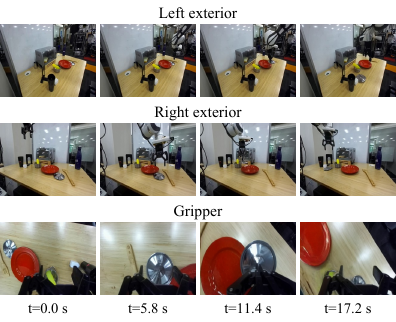}
        \caption{}
    \end{subfigure}\hfill
    \begin{subfigure}[t]{0.48\linewidth}
        \centering
        \includegraphics[width=\linewidth]{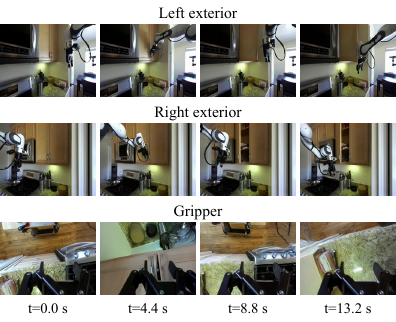}
        \caption{}
    \end{subfigure}\par\vspace{2ex}
    \begin{subfigure}[t]{0.48\linewidth}
        \centering
        \includegraphics[width=\linewidth]{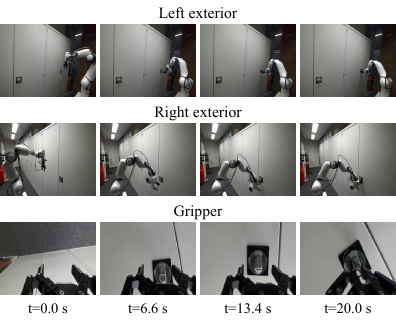}
        \caption{}
    \end{subfigure}\hfill
    \begin{subfigure}[t]{0.48\linewidth}
        \centering
        \includegraphics[width=\linewidth]{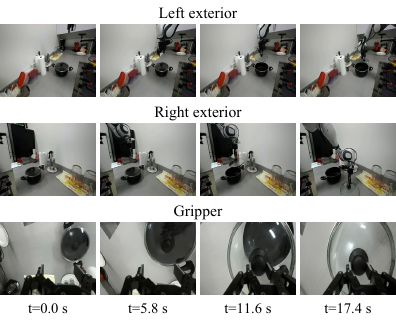}
        \caption{}
    \end{subfigure}
    \caption{\textbf{Cross-embodiment rollouts: DROID.} Six real-world Franka rollouts, tagged (a)--(f), with synchronized left-exterior (top), right-exterior (middle), and gripper (bottom) views, spanning diverse scenes and collection sites. Columns progress forward in time, with elapsed video time in seconds below.}
    \label{fig:ce-droid-all}
\end{figure}

\begin{figure}[t]
    \centering
    \captionsetup[sub]{justification=centering}
    \begin{subfigure}[t]{0.48\linewidth}
        \centering
        \includegraphics[width=\linewidth]{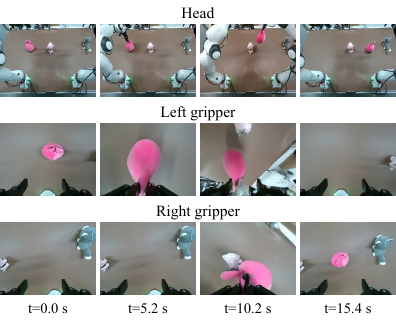}
        \caption{}
    \end{subfigure}\hfill
    \begin{subfigure}[t]{0.48\linewidth}
        \centering
        \includegraphics[width=\linewidth]{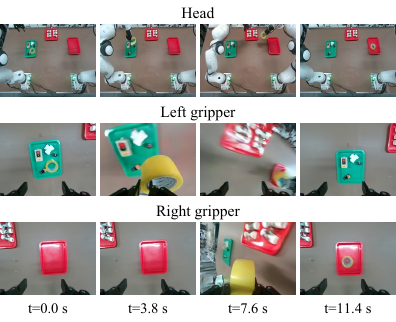}
        \caption{}
    \end{subfigure}\par\vspace{2ex}
    \begin{subfigure}[t]{0.48\linewidth}
        \centering
        \includegraphics[width=\linewidth]{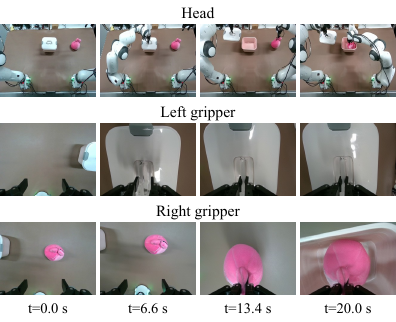}
        \caption{}
    \end{subfigure}\hfill
    \begin{subfigure}[t]{0.48\linewidth}
        \centering
        \includegraphics[width=\linewidth]{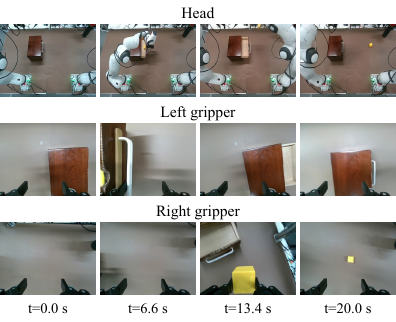}
        \caption{}
    \end{subfigure}\par\vspace{2ex}
    \begin{subfigure}[t]{0.48\linewidth}
        \centering
        \includegraphics[width=\linewidth]{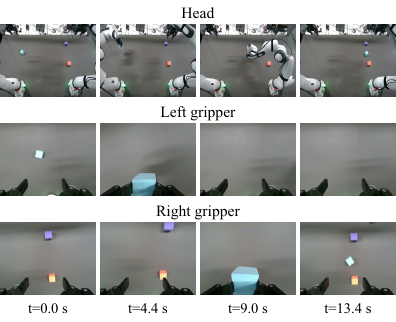}
        \caption{}
    \end{subfigure}\hfill
    \begin{subfigure}[t]{0.48\linewidth}
        \centering
        \includegraphics[width=\linewidth]{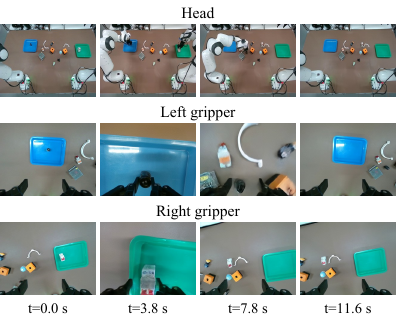}
        \caption{}
    \end{subfigure}
    \caption{\textbf{Cross-embodiment rollouts: RoboMIND~2.0.} Six predicted rollouts, tagged (a)--(f), each pairing synchronized head (top), left-gripper (middle), and right-gripper (bottom) views, covering dual-arm toy handover (a), inter-arm tape passing into a tray (b), open-box toy placement (c), drawer manipulation (d), color-constrained block placement (e), and dual-arm capacitor assembly (f). Columns progress forward in time, with elapsed video time in seconds below.}
    \label{fig:ce-robomind-all}
\end{figure}

\begin{figure}[t]
    \centering
    \captionsetup[sub]{justification=centering}
    \begin{subfigure}[t]{0.48\linewidth}
        \centering
        \includegraphics[width=\linewidth]{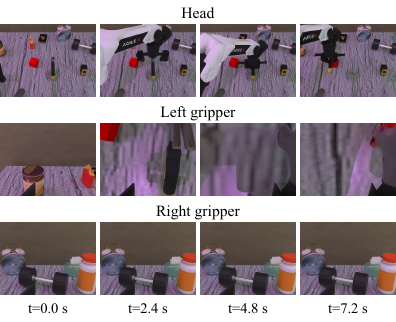}
        \caption{}
    \end{subfigure}\hfill
    \begin{subfigure}[t]{0.48\linewidth}
        \centering
        \includegraphics[width=\linewidth]{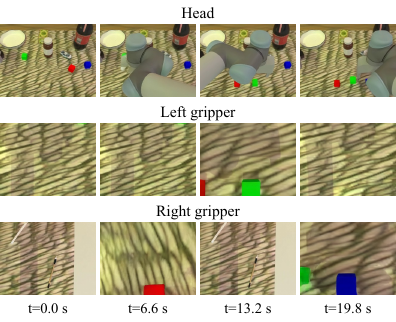}
        \caption{}
    \end{subfigure}\par\vspace{2ex}
    \begin{subfigure}[t]{0.48\linewidth}
        \centering
        \includegraphics[width=\linewidth]{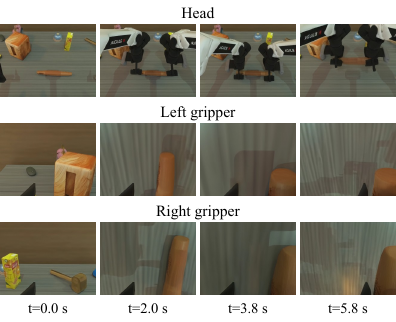}
        \caption{}
    \end{subfigure}\hfill
    \begin{subfigure}[t]{0.48\linewidth}
        \centering
        \includegraphics[width=\linewidth]{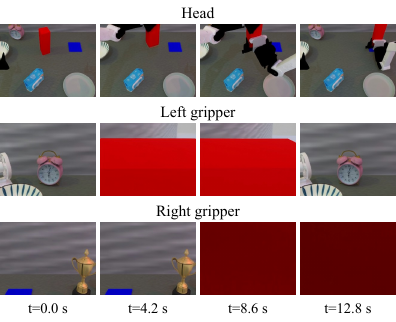}
        \caption{}
    \end{subfigure}\par\vspace{2ex}
    \begin{subfigure}[t]{0.48\linewidth}
        \centering
        \includegraphics[width=\linewidth]{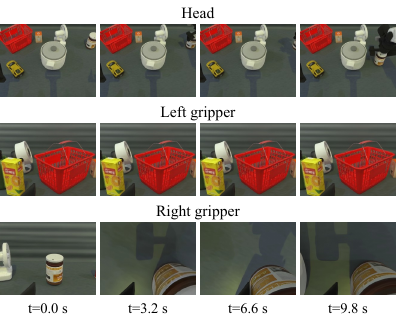}
        \caption{}
    \end{subfigure}\hfill
    \begin{subfigure}[t]{0.48\linewidth}
        \centering
        \includegraphics[width=\linewidth]{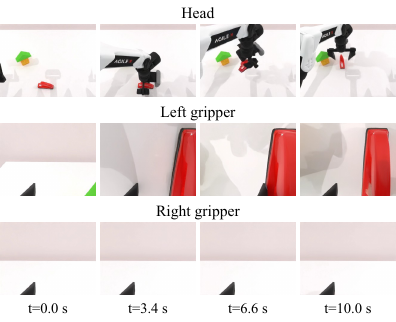}
        \caption{}
    \end{subfigure}
    \caption{\textbf{Cross-embodiment rollouts: RoboTwin~2.0.} Six simulated rollouts, tagged (a)--(f), each pairing synchronized head (top), left-gripper (middle), and right-gripper (bottom) views, covering hammering (a), color ranking (b), roller grasping (c), block handover (d), can-and-pot transport (e), and A-to-B placement (f). Columns progress forward in time, with elapsed video time in seconds below.}
    \label{fig:ce-robotwin-all}
\end{figure}

\Cref{fig:ce-agibot-all,fig:ce-droid-all,fig:ce-robomind-all,fig:ce-robotwin-all} present all twenty-four qualitative clips for the four training benchmarks: six AgiBotWorld-Beta sequences, six DROID sequences, six RoboMIND~2.0 sequences, and six RoboTwin~2.0 sequences. Each example shows four sampled time points, including the first and last video frames. The dataset's camera views are synchronized within each column; $t$ denotes elapsed time in seconds from the start of each prediction video, computed using its frame rate. These clips supplement the representative cross-embodiment qualitative results in the main evaluation.

\clearpage

\clearpage

\clearpage

\end{document}